%% file: ms.tex
\documentclass[sigconf, nonacm]{acmart}

\pdfoutput=1
\newcommand\vldbavailabilityurl{https://github.com/mjschleich/GeCo.jl}

\input{macros}
\usepackage{bm}
\usepackage{pgfplots}
\usepackage{xspace}
\usepackage{balance}

\newcommand{\mysubsubsec}[1]{\noindent\textbf{#1.}}

\usepackage{tikz,tikz-qtree,tikz-qtree-compat}
\usetikzlibrary{fit,topaths,calc,shapes,decorations,chains,scopes,
  arrows,calc,positioning,matrix,backgrounds,shapes.callouts}

\tikzset{
  notice/.style  = { draw, rounded corners=.1cm, rectangle callout, callout relative pointer={#1} },
}

\usepackage{colortbl}

\usepackage{tikzsymbols}
\usetikzlibrary{patterns}
\usepackage{tikzpeople}

\definecolor{oxfordblue}{rgb}{0, 0.33, 0.71}
\definecolor{forestgreen}{rgb}{0.13, 0.55, 0.13}
\definecolor{nicepurple}{rgb}{0.76, 0.14, 0.76}

\colorlet{dred}{red!80!black}
\colorlet{dgreen}{green!50!black}
\colorlet{dorange}{orange!90!black}
\colorlet{dpurple}{nicepurple!80!black}
\colorlet{dmagenta}{magenta!80!black}

\newcommand{\geco}{{\sf GeCo}\xspace}
\newcommand{\ignore}[1]{}

\begin{document}
\title{\geco: Quality Counterfactual Explanations in Real Time}

\author{Maximilian Schleich}
\affiliation{%
  \institution{University of Washington}
}
\email{schleich@cs.washington.edu}

\author{Zixuan Geng}
\affiliation{%
  \institution{University of Washington}
}
\email{zg44@cs.washington.edu}

\author{Yihong Zhang}
\affiliation{%
  \institution{University of Washington}
}
\email{yz489@cs.washington.edu}

\author{Dan Suciu}
\affiliation{%
  \institution{University of Washington}
}
\email{suciu@cs.washington.edu}

\begin{abstract}
  Machine learning is increasingly applied in high-stakes decision
  making that directly affect people’s lives, and this leads to an
  increased demand for systems to explain their decisions.
  Explanations often take the form of {\em counterfactuals}, which
  consists of conveying to the end user what she/he needs to change in
  order to improve the outcome.  Computing counterfactual explanations
  is challenging, because of the inherent tension between a rich
  semantics of the domain, and the need for real time response.  In
  this paper we present \geco, the first system that can compute
  plausible and feasible counterfactual explanations in real time.  At
  its core, \geco\ relies on a genetic algorithm, which is customized
  to favor searching counterfactual explanations with the smallest
  number of changes.  To achieve real-time performance, we introduce
  two novel optimizations: $\Delta$-representation of candidate
  counterfactuals, and partial evaluation of the classifier.  We
  compare empirically \geco against five other systems described in
  the literature, and show that it is the only system that can achieve
  both high quality explanations and real time answers.
\end{abstract}

\maketitle


\ifdefempty{\vldbavailabilityurl}{}{
\vspace{.3cm}
\begingroup\small\noindent\raggedright\textbf{Artifact Availability:}\\
The source code, data, and/or other artifacts have been made available at \url{\vldbavailabilityurl}.
\endgroup
}

\input{include/sec1-intro}
\input{include/sec2-problem}
\input{include/sec3-constraints}

\input{include/sec4-algorithm}

\input{include/sec5-optimizations}

\input{include/sec6-experiments}

\input{include/sec7-conclusions}

\balance

\begin{acks}
  This work was partially supported by NSF IIS 1907997, NSF IIS
  1954222, and a generous gift from RelationalAI.
\end{acks}

\clearpage

\bibliographystyle{ACM-Reference-Format}
\bibliography{main}

\appendix

\input{include/appendix}

\end{document}

%% file: macros.tex
\usepackage{xcolor}

\usepackage{tikz}
\usetikzlibrary{shapes,decorations.markings,arrows.meta,fit,calc,positioning,shapes.gates.logic.US}

\colorlet{LightViolet}{violet!40}
\colorlet{LightRed}{red!40}
\colorlet{LightOrange}{orange!40}
\colorlet{LightGreen}{green!40}
\colorlet{LightBlue}{blue!40}
\colorlet{DarkGreen}{green!50!black}
\colorlet{DarkRed}{red!70!black}
\colorlet{DarkCyan}{red!70!black}
\colorlet{DarkBlue}{blue!80!black}
\definecolor{DarkOrange}{rgb}{1.0, 0.49, 0.0}
\definecolor{Airforceblue}{rgb}{0.36, 0.54, 0.66}

\definecolor{plot1}{HTML}{1f77b4}

\definecolor{tab10_1}{rgb}{0.1216,0.4667,0.7059}
\definecolor{tab10_2}{rgb}{1.0,0.498,0.0549}
\definecolor{tab10_3}{rgb}{0.1725,0.6275,0.1725}
\definecolor{tab10_4}{rgb}{0.8392,0.1529,0.1569}

\definecolor{plot1}{rgb}{0.0,0.6056031611752245,0.9786801175696073}
\definecolor{plot2}{rgb}{0.8888735002725198,0.43564919034818994,0.2781229361419438}
\definecolor{plot3}{rgb}{0.2422242978521988,0.6432750931576305,0.3044486515341153}
\definecolor{plot4}{rgb}{0.7644401754934356,0.4441117794687767,0.8242975359232758}

\newcommand{\defeq}{\stackrel{\mathrm{def}}{=}}

\newcommand{\functionname}[1]{\operatorname{#1}} 

\newcommand{\dom}{\functionname{dom}}

\newcommand{\be}{\begin{enumerate}}
\newcommand{\ee}{\end{enumerate}}
\newcommand{\bi}{\begin{itemize}}
\newcommand{\ei}{\end{itemize}}
\newcommand{\beq}{\begin{equation}}
\newcommand{\eeq}{\end{equation}}

\newcommand{\bp}{\begin{proof}}
\newcommand{\ep}{\end{proof}}
\newcommand{\bcor}{\begin{cor}}
\newcommand{\ecor}{\end{cor}}
\newcommand{\bthm}{\begin{thm}}
\newcommand{\ethm}{\end{thm}}
\newcommand{\blmm}{\begin{lmm}}
\newcommand{\elmm}{\end{lmm}}
\newcommand{\bdefn}{\begin{defn}}
\newcommand{\edefn}{\end{defn}}
\newcommand{\bprop}{\begin{prop}}
\newcommand{\eprop}{\end{prop}}
\newcommand{\bconj}{\begin{conj}}
\newcommand{\econj}{\end{conj}}
\newcommand{\bopm}{\begin{opm}}
\newcommand{\eopm}{\end{opm}}
\newcommand{\bprm}{\begin{prm}}
\newcommand{\eprm}{\end{prm}}
\newcommand{\brmk}{\begin{rmk}}
\newcommand{\ermk}{\end{rmk}}
\newcommand{\bclm}{\begin{claim}}
\newcommand{\eclm}{\end{claim}}
\newcommand{\bex}{\begin{ex}}
\newcommand{\eex}{\end{ex}}

\theoremstyle{plain}            
\newtheorem{thm}{Theorem}
\newtheorem{lmm}[thm]{Lemma}
\newtheorem{prop}[thm]{Proposition}
\newtheorem{cor}[thm]{Corollary}

\theoremstyle{definition}              

\newtheorem{opm}[thm]{Open Problem}
\newtheorem{prm}[thm]{Problem}
\newtheorem{conj}[thm]{Conjecture}
\newtheorem{ex}[thm]{Example}

\newtheorem{defn}{Definition}

\newtheorem{rmk}[thm]{Remark}

\newcounter{claimCounter}
\newtheorem{claim}[claimCounter]{Claim}

\newtheorem*{example*}{Example}

\newtheorem*{lmm*}{Lemma}

\definecolor{Red}{RGB}{255,204,204}
\definecolor{Green}{RGB}{204,255,204}
\definecolor{Blue}{RGB}{204,204,255}

\newcommand{\set}[1]{\{#1\}}                    
\newcommand{\setof}[2]{\{{#1}\mid{#2}\}}        

\usepackage{newfloat}
\DeclareFloatingEnvironment[ fileext=lox, name=Algorithm, placement=tp]{algorithm}

\newcommand{\unioneq}{\ \cup\!=}

\newcommand{\TAB}{\makebox[2.5ex][r]{}}%
\newcommand{\STAB}{\makebox[1.25ex][r]{}}%
\newcommand{\LET}{\textbf{let}\xspace}%
\newcommand{\IF}{\textbf{if}\xspace}%
\newcommand{\ELSE}{\textbf{else}\xspace}%
\newcommand{\UNTIL}{\textbf{until}\xspace}%
\newcommand{\FOREACH}{\textbf{foreach}\xspace}%
\newcommand{\DO}{\textbf{do}\xspace}%
\newcommand{\RETURN}{\textbf{return}\xspace}%

\usepackage{adjustbox}
\usepackage{array}

\newcolumntype{R}[2]{%
    >{\adjustbox{angle=#1,lap=\width-(#2)}\bgroup}%
    l%
    <{\egroup}%
}
\newcommand*\rot{\multicolumn{1}{R{45}{1em}}}


%% file: include/sec1-intro.tex
\section{Introduction}
\label{sec:intro}

Machine learning is increasingly applied in high-stakes decision
making that directly affects people’s lives. As a result, there is a
huge need to ensure that the models and their predictions are
interpretable by their human users.  Motivated by this need, there has
been a lot of recent interest within the machine learning community in
techniques that can explain the outcomes of models. Explanations
improve the transparency and interpretability of the underlying model,
they increase user's trust in the model predictions, and they are a
key facilitator to evaluate the fairness of the model for
underrepresented demographics.  The ability to explain is no longer a
nice-to-have feature, but is increasingly required by law; for
example, the GDPR regulations grant users the \emph{right to
  explanation} to automated decision
algorithms~\cite{wachter2017counterfactual}.  In addition to
supporting the end user, explanations can also be used by model
developers to debug and monitor their ever more complex models.

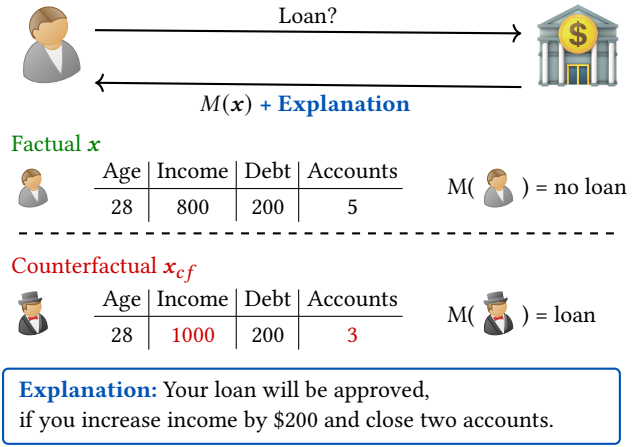
\begin{figure}
  \begin{center}

    \begin{tikzpicture}
      \tikzset{data/.style={draw, rectangle, rounded corners = .07cm, align=center, inner sep = .2cm, outer sep = .1 cm}}

      \node[bob,scale=2] (joe) at (0,0) {};
      \node[inner sep=0pt] (bank) at (7,0) {\includegraphics[width=1.1cm]{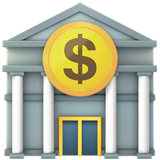}};

      \draw[->, thick] ($(joe.north east)+(0.2,-0.3)$) -- node [above,midway] {Loan?} ($(bank.north west)+(-0.2,-0.4)$);
      \draw[<-, thick] ($(joe.south east)+(0.2,0)$) --  node [below,midway] {$M(\bm x)$ {\color{oxfordblue}\bf + Explanation }} ($(bank.south west)+(-0.2,0)$);

    \end{tikzpicture}
  \begin{tikzpicture}
    \tikzset{data/.style={draw, rectangle, rounded corners = .07cm, align=center, inner sep = .2cm, outer sep = .1 cm}}

    \node[anchor=west, dgreen] at (-0.4,.6) {Factual $\bm x$};
    \node[bob] (bob) at (0,0) {};

    \matrix (m) at ($(bob.east)+(2.7,0)$)
    [matrix of nodes, nodes in empty cells, ampersand replacement=\&]
    {
      Age  \& Income \& Debt \& Accounts \\
      28   \& 800    \& 200  \& 5        \\
    };

    \draw ($(m-1-1.south west)+(0,0.05)$)-- ++(4.1,0); 
    \draw ($(m-1-2.north west)+(0,-0.1)$) -- ++(0,-.8);
    \draw ($(m-1-3.north west)+(0,-0.1)$) -- ++(0,-.8);
    \draw ($(m-1-4.north west)+(0,-0.1)$) -- ++(0,-.8);

    \node[bob] (cbob) at ($(bob.east)+(6,0)$) {};
    \node at ($(cbob)+(-0.5,0)$) {M(};
    \node[anchor=west] at ($(cbob)+(0.2,0)$) {) = no loan};

    \draw[thick,dashed] ($(bob.south west)+(0,-0.35)$) -- ++(8,0);


    \node[anchor=north west, dred] at (-.4,-.8) {Counterfactual $\bm x_{cf}$};
    \node[groom] (richbob) at (0,-1.7) {};

    \matrix (m2) at ($(richbob.east)+(2.7,0)$)
    [matrix of nodes, nodes in empty cells, ampersand replacement=\&]
    {
      Age \& Income \& Debt \& Accounts                                                              \\
      28  \& {\color{dred}1000} \& 200 \& {\color{dred}3} \\
    };

    \draw ($(m2-1-1.south west)+(0,0.05)$) -- ++(4.1,0); 
    \draw ($(m2-1-2.north west)+(0,-0.1)$) -- ++(0,-.8);
    \draw ($(m2-1-3.north west)+(0,-0.1)$) -- ++(0,-.8);
    \draw ($(m2-1-4.north west)+(0,-0.1)$) -- ++(0,-.8);

    \node[groom] (crbob) at ($(richbob.east)+(6,0)$) {};
    \node at ($(crbob)+(-0.5,0)$) {M(};
    \node[anchor=west] at ($(crbob)+(0.2,0)$) {) = loan};

    \node[anchor=west] at (-0.3,-2.7) {{\color{oxfordblue}\bf Explanation:} Your loan will be approved, };
    \node[anchor=west] at (-0.3,-3.1) {if you increase income by \$200 and close two accounts.};
    \draw[thick, oxfordblue, rounded corners=.07cm] (-0.4,-2.4) -- (7.8,-2.4) -- (7.8,-3.4) -- (-0.4,-3.4) -- cycle;
  \end{tikzpicture}
\end{center}

  \caption{Example of a counterfactual explanation
    scenario.}
  \label{fig:example}
\end{figure}

In this paper we focus on \emph{local explanations}, which provide post-hoc
explanations for one single prediction, and are in contrast to \emph{global
explanations}, which aim to explain the entire model (e.g. for debugging
purposes).  In particular we study \emph{counterfactual explanations}: given an
instance $\bm x$, on which the machine learning model predicts a negative,
``bad'' outcome, the explanation says what needs to change in order to get the
positive, ``good'' outcome, usually represented by a counterfactual example $\bm
x_{cf}$.  For example, a customer applies for a loan with a bank, the bank
denies the loan application, and the customer asks for an explanation; the
system responds by indicating what features need to change in order for the loan
to be approved, see Fig.~\ref{fig:example}.  In the AI literature,
counterfactual explanations are currently considered the most attractive types
of local explanations, even for models that are considered ``interpretable'',
such as random forests or generalized linear
models~\cite{mace,wachter2017counterfactual,ustun:recourse,wexler2019wit},
because they offer actionable feedback to the customers, and have been deemed
satisfactory by some legislative bodies, for example they are deemed to satisfy
the GDPR requirements~\cite{wachter2017counterfactual}.

The major challenge in computing a counterfactual explanation is the tension
between a rich semantics on the one hand, and the need for real-time, interactive
feedback on the other hand.  The semantics needs to be rich in order to reflect
the complexities of the real world. We want $\bm x_{cf}$ to be as close as
possible to $\bm x$, but we also want $\bm x_{cf}$ to be {\em plausible},
meaning that its features should make sense in the real world.  We also want the
transition from $\bm x$ to $\bm x_{cf}$ to be {\em feasible}, for example {\em
    age} should only increase.  The plausibility and feasibility constraints are
dictated by laws, societal norms, application-specific requirements, and may
even change over time; an explanation system must be able to support constraints
with a rich semantics.  Moreover, the search space for counterfactuals is huge,
because there are often hundreds of features, and each can take values from some
large domain.  On the other hand, the computation of counterfactuals needs to be
done at interactive speed, because the explanation system is eventually
incorporated in a user interface. Performance has been identified as the main
challenge for deployment of counterfactual explanations in
industry~\cite{bhatt2020explainable,wexler2019wit}.
\ignore{
}
The tension between performance and rich semantics is the main technical
challenge in counterfactual explanations.  Previous systems either explore a
complete search space with rich semantics, or answer at interactive speed, but
not both: see Fig.~\ref{fig:taxonomy} and discussions in Section~\ref{sec:exp}.
For example, on one extreme MACE~\cite{mace} enforces plausibility and
feasibility by using a general-purpose constraint language, but the solver often
takes many minutes to find a counterfactual. At the other extreme, Google's
What-if Tool (WIT)~\cite{wexler2019wit} restricts the search space to a fixed
dataset of examples, ensuring fast response time, but poor explanations.

\begin{figure}[t]
  \centering
  \begin{tabular}{l|l|l|}
    \multicolumn{1}{l}{} & \multicolumn{1}{l}{Limited}                                  & \multicolumn{1}{l}{Complete}     \\
    \multicolumn{1}{l}{} & \multicolumn{1}{l}{search space}                             & \multicolumn{1}{l}{search space} \\ \cline{2-3}
                         &                                                              &                                  \\[-0.6em]
    Non-interactive      & CERTIFAI~\cite{certifai}                                     & MACE~\cite{mace}                 \\[-0.6em]
                         &                                                              &                                  \\ \cline{2-3}
                         &                                                              &                                  \\[-0.6em]
    Interactive          & What-If~\cite{wexler2019wit} DiCE~\cite{mahajan2019causalcf} & \geco\ (this paper)              \\[-0.6em]
                         &                                                              &                                  \\ \cline{2-3}
  \end{tabular}
  \caption{Taxonomy of Counterfactual Explanation Systems}
  \label{fig:taxonomy}
\end{figure}

In this paper we present \geco, the first interactive system for
counterfactual explanations that supports a complex, real-life
semantics of counterfactuals, yet provides answers in real
time.
At its core, \geco\ defines a search space of counterfactuals using a
plausibility-feasibility constraint language, PLAF, and a database $D$.  PLAF is
used to define constraints like ``{\em age} can only increase'', while the
database $D$ is used to capture correlations, like ``{\em job-title} and {\em
salary} are correlated''.  By design, the search space of possible
counterfactuals is huge.  To search this space, we make a simple observation. A
good explanation $\bm x_{cf}$ should differ from $\bm x$ by only a few features;
counterfactual examples $\bm x_{cf}$ that require the customer to change too
many features are of little interest.  Based on this observation, we propose a
{\em genetic algorithm}, which we customize to search the space of
counterfactuals by prioritizing those that have fewer changes.  Starting from a
population consisting of just the given entity $\bm x$, the algorithm repeatedly
updates the population by applying the operations {\em crossover} and {\em
mutation}, and then {\em selecting} the best counterfactuals for the new
generation.  It stops when it reaches a sufficient number of examples on which
the classifier returns the ``good'' (desired) outcome. While counterfactual
explanations can be applied to various data types, e.g. DiCE can explain image
classifications~\cite{mahajan2019causalcf}, we focus on structured tabular data.
Structured data is the predominant data type used in financial services, a key
target industry for counterfactual explanations. We also assume that the model
and data is static; the impact of updates on the explanations is an challenging
direction for future work.

The main performance limitation in \geco\ is its innermost loop.  By the nature
of the genetic algorithm, \geco\ needs to repeatedly add and remove
counterfactuals to and from the current population, and ends up having to
examine thousands of candidates, and apply the classifier $M(\bm x')$ on each of
them.  We propose two novel optimizations to speedup the inner loop of \geco:
$\Delta$-representation and classifier specialization via partial evaluation. In
  {\em $\Delta$-representation} we group the current population by the set of
features $\Delta F$ by which they differ from $\bm x$, and represent this entire
subpopulation in a single relation whose attributes are only $\Delta F$; for
example all candidate examples $\bm x'$ that differ from $\bm x$ only by {\em
    age} are represented by a single-column table, storing only the modified {\em
    age}.  This leads to huge memory savings over the naive representation (storing
all features of all candidates $\bm x'$), which, in turn, leads to performance
improvements.  {\em Partial evaluation} specializes the code of the classifier
$M$ to entities $\bm x'$ that differ from $\bm x$ only in $\Delta F$; for
example, if $M$ is a decision tree, then normally $M(\bm x')$ inspects all
features from the root to a leaf, but if $\Delta F$ is {\em age}, then after
partial evaluation it only needs to inspect {\em age}, for example $M$ can be
$\textit{age} > 30$, or perhaps $30 < \textit{age} < 60$, because all the other
features are constants within this subpopulation. $\Delta$-representation and
partial evaluation work well together, and, when combined, allows \geco\ to
compute the counterfactual in interactive time. At a high level, our
optimizations belong to a more general effort that uses database and program
optimizations to improve the performance of machine learning tasks (e.g.,
\cite{lmfao, OS:SIGREC:2016, kumar2015learning, 10.14778/3317315.3317323, spark-sql}).

We benchmarked \geco against four counterfactual explanation systems:
MACE~\cite{mace}, DiCE~\cite{mahajan2019causalcf}, What-If
Tool~\cite{wexler2019wit}, and CERTIFAI~\cite{certifai}. We show that in all
cases \geco\ produced the best quality explanation (using quality metrics
defined in~\cite{mace}), at interactive speed.  We also conduct
micro-experiments showing that the two optimizations, $\Delta$-representation
and partial evaluation, account for a performance improvement of up to
$5\times$, and thus are critical for an interactive deployment of \geco.


{\bf Discussion}
Explanation techniques can be broadly categorized as {\em white-box} or {\em
    black-box}. {\em White-box} explanations are designed for a specific class of
models and domains, e.g., Krpyton~\cite{10.1145/3299869.3319874} is specific to
neural networks for image classification. These explanations exploit the
structural properties of the classifier and typically cannot be applied across
domains. While there is no commonly agreed definition of {\em black-box}
explanations, we adopt a strict definition in this paper: the explanation is
  {\em black-box} if the classifier $M$ is available only as an oracle that, when
given an input $\bm x'$, returns the outcome $M(\bm x')$. Thus, black-box
explanations do not enforce any restrictions on the underlying domain and
classifier. This is particularly useful when the model is subject to IP
restrictions or provided through an API, e.g., the Google Vision API.
A black-box classifier makes it difficult for the system to compute an
explanation that is {\em consistent} with the underlying classifier, meaning
that $M(\bm x_{cf})$ returns the desired, ``good'', outcome.  As an example,
LIME~\cite{lime} uses a black-box classifier, learns a simple, interpretable
model locally, around the input data point $\bm x$, and uses it to obtain an
explanation; however, its explanations are not always consistent with the
predictions of the original classifier~\cite{rudin2019stop, slack2020fooling}.

A key advantage of counterfactual explanations is that they guarantee model
consistency and they can be black-box. A black-box model, however, makes it
difficult to explore the search space of counterfactuals, because we cannot
examine the code of $M$ for hints of how to quickly get from $\bm x$ to a
counterfactual $\bm x_{cf}$. For this reason, several previous systems claim to
be black box, but require access to the code of the classifier. We classify
these systems as {\em gray-box}: A {\em gray-box} explanation has access to the
code of the model $M$. Early counterfactual explanation systems are based
on gradient descent~\cite{wachter2017counterfactual, mothilal2020diverse}. They
are gray-box, because they need access to the code of $M$ in order to compute
its gradient. MACE~\cite{mace} is also gray-box, since it translates both the
classifier logic and the feasibility/plausibility constraints into a logical
formula and then solves for counterfactuals via multiple calls to an SMT solver.
These systems have similar restrictions as white-box explanations. In contrast,
CERTIFAI~\cite{certifai} and Google's What-if Tool (WIT)~\cite{wexler2019wit}
are fully black-box, but limit the search space (cf. Fig.~\ref{fig:taxonomy}).
CERTIFAI is based on a genetic algorithm, but the quality of its explanations
are highly sensitive to the computation of the initial population; we discuss
CERTIFAI in detail in Sec.~\ref{sec:conv-ga}. WIT severely limits its search
space to a given dataset.

Our system, \geco, is designed to work with any black-box model and to explore
the large search space of counterfactuals. Yet, if the code of the classifier is
available, then we use the partial evaluation optimization to improve the
runtime. Thus, \geco differs from other systems in that it uses access to the
code of $M$ not to guide the search, but only to optimize the execution time.

Counterfactual explanations are related to the problem of finding adversarial
examples~\cite{szegedy2014adversarial, carlini2017adversarial, guo2019simba,
Madry2018adversarial}, which are small perturbations of the input instance that
lead to a different classification. Such  examples are used to evaluate the
robustness of the classifier. A key difference is that adversarial examples
typically have many, almost indistinguishable changes, which are not required to
be plausible or feasible. In Sec.~\ref{sec:exp}, we show that techniques for
adversarial examples are not directly applicable to the problem of finding
quality counterfactual explanations by comparing GeCo with an adaptation of the
SimBA algorithm~\cite{guo2019simba}.

{\bf Contributions} In summary, our contributions are as follows.
\begin{itemize}
  \item We describe \geco, the first system that computes feasible and plausible
        explanations in interactive response time.
  \item We describe the search space of \geco\ consisting of a database
        $D$ and a constraint language PLAF. Section~\ref{sec:constraints}.
  \item We describe a custom genetic algorithm for exploring the space
        of candidate counterfactuals.  Section~\ref{sec:algorithm}.
  \item We describe two optimization techniques: $\Delta$-representation
        and partial evaluation. Section~\ref{sec:optim}.
  \item We conduct an extensive experimental evaluation of \geco, and compare it
        to MACE, Google's What-If Tool, CERTIFAI, and an adaptation of SimBA.
        Section~\ref{sec:exp}.
\end{itemize}

%% file: include/sec2-problem.tex
\newcommand{\desired}{\texttt{``good''}}

\section{Counterfactual Explanations}
\label{sec:problem}

We consider $n$ features $F_1, \ldots, F_n$, with domains
$\dom(F_1), \ldots, \dom(F_n)$.  We assume to have a black-box model
$M$, i.e. an oracle that, when given a feature vector
$\bm x = (x_1, \ldots, x_n)$, returns a prediction $M(\bm x)$.  The
prediction is a number between 0 and 1, where 1 is the desired, or
good outcome, and 0 is the undesired outcome.  For simplicity, we will
assume that $M(\bm x) > 0.5$ is ``good'', and everything else is
``bad''.  If the classifier is categorical, then we simply replace its
outcomes with the values $\set{0,1}$.  Given an instance $\bm x$ for
which $M(\bm x)$ is ``bad'', the goal of the counterfactual
explanation is to find a counterfactual instance $\bm x_{cf}$ such
that (1) $M(\bm x_{cf})$ is ``good'', (2) $x_{cf}$ is close to $x$,
and (3) $\bm x_{cf}$ is both {\em feasible} and {\em
  plausible}. Formally, a counterfactual explanation is a solution to
the following optimization problem:
\begin{align}
  \arg\min_{\bm x_{cf}}&\  \text{dist}(\bm x,\bm x_{cf}) \label{eq:prob}\\
  s.t.&\  M(\bm x_{cf}) > 0.5 \nonumber \\
      & \bm x_{cf} \in \mathcal{P} && \texttt{// $x_{cf}$ is plausible} \nonumber\\
      & \bm x_{cf} \in \mathcal{F}(\bm x) && \texttt{// $x_{cf}$ is feasible}  \nonumber
\end{align}
where $\text{dist}$ is a distance function.  The counterfactual
explanation $\bm x_{cf}$ ranges over the space
$\dom(F_1) \times \cdots \times \dom(F_n)$, subject to the
plausibility and feasibility constraints $\mathcal{P}$ and
$\mathcal{F}(\bm x)$, which we discuss in the next section.  In
practice, as we shall explain, \geco\ returns not just one, but the
top $k$ best counterfactuals $\bm x_{cf}$.

The role of the distance function is to ensure that \geco\ finds the
\emph{nearest counterfactual} instance that satisfies the
constraints. In particular, we are interested in counterfactuals that
change the values of only a few features, which helps define concrete
actions that the user can perform to achieve the desired outcome.  For
that purpose, we use the distance function $\text{dist}(\bm x, \bm y)$
introduced by MACE~\cite{mace}, which we briefly review here.

We start by defining domain-specific distance functions
$\delta_1, \ldots, \delta_n$ for each of the $n$ features, as follows.
If $\dom(F_i)$ is categorical, then $\delta_i(x_i, y_i) \defeq 0$ if
$x_i=y_i$ and $\delta_i(x_i, y_i) \defeq 1$ otherwise.  If $\dom(F_i)$
is a continuous or an ordinal domain, then
$\delta_i(x_i,y_i) \defeq |x_i - y_i| / w$, where $w$ is the range of
the domain.  We note that, when the range is unbounded, or unknown,
then alternative normalizations are possible, such as the Median
Absolute Distance (MAD)~\cite{wachter2017counterfactual}, or the
standard deviation~\cite{wexler2019wit}.
We define the $\ell_p$-distance between $\bm x, \bm y$ as:
\begin{align*}
  \text{dist}_p(\bm x, \bm y) \defeq & \left(\sum_i \delta_i^p(x_i,y_i)\right)^{1/p}
\end{align*}
and adopt the usual convention that the $\ell_0$-distance is the
number of distinct features: $\text{dist}_0(\bm x, \bm y) \defeq
|\setof{i}{\delta_i(x_i,y_i)\neq 0}|$. Finally, we define our distance
function as a weighted combination of the $\ell_0, \ell_1$, and
$\ell_\infty$ distances:
\begin{align}
  dist(\bm x, \bm y) = & \alpha \cdot \frac{\text{dist}_0(\bm x, \bm y)}{n} + \beta \cdot \frac{\text{dist}_1(\bm x, \bm y)}{n} +  \gamma \cdot \text{dist}_\infty(\bm x, \bm y)
\label{eq:dist}
\end{align}
where $\alpha, \beta, \gamma \geq 0$ are hyperparameters, which must
satisfy $\alpha + \beta +\gamma = 1$.  Notice that
$0 \leq dist(\bm x, \bm y) \leq 1$.  The intuition is that the
$\ell_0$-norm restricts the number of features that are changed, the
$\ell_1$ norm accounts for the average change of distance between
$\bm x$ and $\bm y$, and the $\ell_\infty$ norm restricts the maximum
change across all features.  Although~\cite{mace} defines the distance
function~\eqref{eq:dist}, the MACE system hardwires the
hyperparameters to $\alpha=0, \beta=1, \gamma=0$; we discuss this, and
the setting of different hyperparameters in
Sec.~\ref{sec:exp}.

%% file: include/sec3-constraints.tex
\section{The Search Space}
\label{sec:constraints}

The key to computing high quality explanations is to define a complete
space of candidates that the system can explore.  In \geco\ the search
space for the counterfactual explanations is defined by two
components: a database of entities
$D = \set{\bm x_1, \bm x_2, \ldots}$ and plausibility and feasibility
constraint language called PLAF.

The database $D$ can be the training set, a test set, or
historical data of past customers for which the system has performed
predictions.  It is used in two ways.  First, \geco computes the
domain of each feature as the active domain in $D$:
$\dom(F_i) \defeq \Pi_{F_i}(D)$.  Second, the data analyst can specify
{\em groups} of features with the command:
\begin{align*}
  & \texttt{GROUP } F_{i_1}, F_{i_2}, \ldots
\end{align*}
and in that case the joint domain of these features is restricted to
the combination of values found in $D$,
i.e. $\Pi_{F_{i_1}F_{i_2}\cdots}(D)$.  Grouping is useful in several
contexts.  The first is when the attributes are correlated.  For
example, if we have a functional dependency like
$\texttt{zip}\rightarrow \texttt{city}$, then the data analyst would
group $\texttt{zip},\texttt{city}$.  For another example of
correlation, consider $\texttt{education}$ and $\texttt{income}$.
They are correlated without satisfying a strict functional dependency;
by grouping them, the data analyst ensures that \geco\ considers only
combinations of values found in the dataset $D$.  As a final example,
consider attributes that are the result of one-hot encoding:
e.g. $\texttt{color}$ may be expanded into
$\texttt{color\_red}, \texttt{color\_green}, \texttt{color\_blue}$.
By grouping them together, the analyst restricts \geco\ to consider
only values $(1,0,0), (0,1,0), (0,0,1)$ that actually occur in $D$.

The constraint language PLAF allows the data analyst to specify which
combination of features of $\bm x_{cf}$ are plausible, and which can
be feasibly reached from $\bm x$.  PLAF consists of statements of the
form:
\begin{align}
\texttt{PLAF IF }  \Phi_1 \texttt{ and } \Phi_2 \texttt{ and } \cdots \texttt{ THEN } & \Phi_0
\label{eq:plaf:def}
\end{align}
where each $\Phi_i$ is an atomic predicate of the form
$e_1 \text{ op } e_2$ for
$\text{op} \in \set{=, \neq, \leq, <, \geq , >}$, and each expression
$e_1, e_2$ is over the {\em current} features, denoted
$\texttt{x}.F_i$, and/or {\em counterfactual} features, denoted
$\texttt{x\_cf}.F_i$.  The \texttt{IF} part may be missing.

\begin{example} \label{ex:plaf}
  Consider the following PLAF specification:
  \begin{align}
    & \texttt{GROUP education, income} \label{eq:plaf:0} \\
    & \texttt{PLAF x\_cf.gender $=$ x.gender} \label{eq:plaf:1} \\
    & \texttt{PLAF x\_cf.age $>=$ x.age} \label{eq:plaf:2} \\
    & \texttt{PLAF IF x\_cf.education > x.education} \nonumber \\
    & \ \ \ \ \ \ \ \ \texttt{THEN x\_cf.age > x.age+4}\label{eq:plaf:3}
  \end{align}
  The first statement says that \texttt{education} and \texttt{income}
  are correlated: \geco\ will consider only counterfactual values that
  occur together in the data.  Rule~\eqref{eq:plaf:1} says that
  \texttt{gender} cannot change, rule~\eqref{eq:plaf:2} says that
  \texttt{age} can only increase, while rule~\eqref{eq:plaf:3} says
  that, if we ask the customer to get a higher education degree, then
  we should also increase the \texttt{age} by 4.  The last
  rule~\eqref{eq:plaf:3} is adapted from~\cite{mahajan2019causalcf},
  who have argued for the need to restrict counterfactuals to those
  that satisfy {\em causal} constraints.
\end{example}

PLAF has the following restrictions.  (1) The groups have to be
disjoint: if a feature $F_i$ needs to be part of two groups then they
need to be union-ed into a larger group.  We denote by $G(F_i)$ the
unique group containing $F_i$ (or $G(F_i)=\set{F_i}$ if $F_i$ is not
explicitly included in any group).  (2) the rules must be acyclic, in
the following sense.  Every consequent $\Phi_0$ in
~\eqref{eq:plaf:def} must be of the form $\texttt{x\_cf.$F_i$ op } e$,
i.e. must ``define'' a counterfactual feature $F_i$, and the following
graph must be acyclic: the nodes are the groups
$G(F_1), G(F_2), \ldots$, and the edges are $(G(F_j),G(F_i))$ whenever
there is a rule~\eqref{eq:plaf:def} that defines $\texttt{x\_cf}.F_i$
and that also contains $\texttt{x\_cf}.F_j$.  We briefly illustrate
with Example~\ref{ex:plaf}.  The three
rules~\eqref{eq:plaf:1}-\eqref{eq:plaf:3} ``define'' the features
$\texttt{gender}, \texttt{age}$, and $\texttt{age}$ respectively, and
there there is a single edge
$\set{\texttt{education, income}} \rightarrow \set{\texttt{age}}$,
resulting from Rule~\eqref{eq:plaf:3}.  Therefore, the PLAF program is
acyclic.

The restrictions are not limiting the expressive power of PLAF, but only
encourage users to write constraints that \geco\ can evaluate efficiently.
Indeed, consider any CNF formula $C_1 \wedge C_2 \wedge \cdots$ where each
clause $C_i$ has the form $\Phi_1 \vee \Phi_2 \vee \cdots$ If $\Phi_i'$ is the
negation of $\Phi_i$ (e.g. the negation of $e_1 \leq e_2$ is $e_1 > e_2$), then
we can write the clause as the PLAF statement $\Phi_1' \wedge \Phi_2' \wedge
\cdots \Rightarrow (1=2)$, where $1=2$ stands for \texttt{false}.  The PLAF
program is acyclic because the precedence graph has no edges, but it would force
\geco\ to search for counterfactuals through rejection sampling only. Instead,
by encouraging users to write constraints where each counterfactual feature is
defined by a rule as explained above, we reduce the need for, or completely
avoid rejection sampling.

%% file: include/sec4-algorithm.tex
\newcommand{\pop}{\ensuremath{\texttt{POP}}\xspace}
\newcommand{\popext}{\ensuremath{\texttt{POPEXT}}\xspace}
\newcommand{\topk}{\ensuremath{\texttt{TOPK}}\xspace}
\newcommand{\dcand}{\ensuremath{\texttt{CAND}}\xspace}
\newcommand{\xcf}{\ensuremath{\bm x_{cf}}\xspace}

\begin{algorithm}[t]\centering
  \begin{tabular}{|p{0.95\columnwidth}|}\hline
    {\bf explain} (instance $\bm x$, $\text{classifier } M$,  $\text{dataset } D$, $\text{PLAF } (\Gamma, \texttt{C})$)\\\hline\\[-0.8em]
    $\texttt{C}_{\bm x} = \textbf{ground}(\bm x, \texttt{C}); \TAB \texttt{DG} = \textbf{feasibleSpace}(D, \Gamma, \texttt{C}_{\bm x});$ \\
    $\pop = [\; (\bm x, \emptyset) \;]$  \\
    $\pop = \textbf{mutate}(\pop, \Gamma, \texttt{DG}, \texttt{C}_{\bm x}, m_{\text{init}}$)  \hfill{\small\texttt{//initial population}} \\
    $\pop = \textbf{selectFittest}(\bm x, \pop, M, q)$\\[0.2em]
    $\DO\STAB\{$\\
    $\TAB \dcand = \textbf{crossover}(\pop, \texttt{C}_{\bm x})\cup \textbf{mutate}(\pop,\Gamma, \texttt{DG}, \texttt{C}_{\bm x}, m_{\text{mut}})$\\[0.2em]
    $\TAB \pop = \textbf{selectFittest}(\bm x, \pop \cup \dcand, M, q)$\\[0.2em]
    $\TAB \topk = \pop[1:k]$\\[0.2em]
    $\}\STAB \UNTIL\ (\ \textbf{counterfactuals}(\topk, M) \textbf{ and }  \topk \cap \dcand = \emptyset \ )$\\[0.2em]
    $\RETURN\ \topk$\\\hline
  \end{tabular}
  \caption{Pseudo-code of \geco's custom genetic algorithm to generate
    counterfactual explanations.}
\label{alg:genalg}
\end{algorithm}

\section{\geco's Custom Genetic Algorithm}
\label{sec:algorithm}

In this section, we introduce the custom genetic algorithm that \geco
uses to efficiently explore the space of counterfactual
explanations. A genetic algorithm is a meta-heuristic for constraint
optimization that is based on the process of natural selection. There
are four core operations. First, it defines an \textbf{initial
  population} of candidates. Then, it iteratively selects the
\textbf{fittest} candidates in the population, and generates new
candidates via \textbf{mutate} and \textbf{crossover} on the selected
candidates, until convergence.

While there are many optimization algorithms that could be used to
solve our optimization problem (defined in Eq~\eqref{eq:prob}), we
chose a genetic algorithm for the following reasons: (1) The genetic
algorithm is easily customizable to the problem of finding
counterfactual explanations; (2) it seamlessly supports the rich
semantics of PLAF constraints, which are necessary to ensure that the
explanations are feasible and plausible; (3) it does not require any
restrictions on the underlying classifier and data, and thus is able
to provide black-box explanations; and (4) it returns a diverse set of
explanations, which may provide different actions that can
lead to the desired outcome.

In \geco, we customized the core operations of the genetic algorithm based on
the following key observation:  A good explanation $\bm x_{cf}$ should differ
from $\bm x$ by only a few features; counterfactual examples $\bm x_{cf}$ that
require the customer to change too many features are of little interest. For
this reason, \geco first explores counterfactuals that change only a single
feature group, before exploring increasingly more complex action spaces in
subsequent generations.

In the following, we first overview of the genetic algorithm used by \geco and
then provide additional details for the core operations.

\subsection{Overview}

\geco's pseudocode is shown in Algorithm~\ref{alg:genalg}.  The inputs
are: an instance $\bm x$, the black-box classifier $M$, a dataset $D$,
and PLAF program $(\Gamma, C)$.  Here $\Gamma=\set{G_1, G_2, \ldots}$
are the groups of features, and $C = \set{C_1, C_2, \ldots}$ are the
PLAF constraints, see Sec.~\ref{sec:constraints}. The algorithm has
four integer hyperparameters $k,m_{\text{init}},m_{\text{mut}},q>0$,
with the following meaning: $k$ represents the number of
counterfactuals that the algorithm returns to the user; $q$ is the
size of the population that is retained from one generation to the
next; and $m_{\text{init}}$, $m_{\text{mut}}$ control the number of
candidates that are generated for the initial population, and during
mutation respectively.  We always set $k < q$.

As explained in Sec.~\ref{sec:constraints}, the active domain of each
attribute are values found in the database $D$.  More generally, for
each group $G_i \in \Gamma$, its values must be sampled together from
those in the database $D$.  The \geco\ algorithm starts by grounding
(specializing) the PLAF program $C$ to the entity $\bm x$ (the
$\textbf{ground}$ function), then calls the \textbf{feasibleSpace}
operator, which computes for each group $G_i$ a relation $DG_i$ with
attributes $G_i$ representing the sample space for the group $G_i$; we
give details in Sec.~\ref{sec:sample-space}.

Next, \geco\ computes the initial population.  In our customized
algorithm, the initial population is obtained simply by applying the
mutate operator to the given entity $\bm x$.  Throughout the execution
of the algorithm, the population is a set of pairs
$(\bm x', \Delta')$, where $\bm x'$ is an entity and $\Delta'$ is the
set of features that were changed from $\bm x$.  Throughout this
section we call the entities $\bm x'$ in the population {\em
  candidates}, or {\em examples}, but we don't call them
counterfactuals, unless $M(\bm x')$ is ``good'', i.e.
$M(\bm x') > 0.5$ (see Sec.~\ref{sec:problem}).  In fact, it is
possible that none of the candidates in the initial population are
classified as good.  The goal of \geco is to find at least $k$
counterfactuals through a sequence of mutation and crossover
operation.

The main loop of \geco's genetic algorithm consists of extending the population
with new candidates obtained by mutation and crossover, then keeping only the
$q$ fittest for the next generation.  The operators \textbf{selectFittest},
\textbf{mutate}, \textbf{crossover} are described in Sec.~\ref{sec:selection},
~\ref{sec:mutation}, and~\ref{sec:crossover}.  The algorithm stops when the top
$k$ (from the select $q$ fittest) candidates are all counterfactuals {\em and}
are stable from one generation to the next; the function
\textbf{counterfactuals} simply tests that all candidates are counterfactuals,
by checking\footnote{In our implementation we store the value $M(\bm x')$
together with $\bm x'$, so we don't have to compute it repeatedly.  We omit some
details for simplicity of the presentation.} that $M(\bm x')$ is ``good''. We
describe now the details of the algorithm.

\subsection{Feasible Space Operators}
\label{sec:sample-space}

Throughout the execution of the genetic algorithm, \geco ensures that
all candidates $\bm x'$ satisfy all PLAF constraints $C$.  It achieves
this efficiently through three functions: $\textbf{ground}$,
$\textbf{feasibleSpace}$, and $\textbf{actionCascade}$.  We describe
these functions here, and omit their pseudocode (which is
straightforward).

The function $\textbf{ground}(\bm x, C)$ simply instantiates all
features of $\bm x$ with constants, and returns ``grounded'' constraints
$C_{\bm x}$.  All candidates $\bm x'$ will need to satisfy these
grounded constraints.
\begin{example}\label{ex:plaf-grounded}  Let $C$ be the three
  constraints of the PLAF program in Example~\ref{ex:plaf}.  Assume
  that the instance is:
  \begin{align*}
  \bm x = & (\texttt{gender}=\texttt{female}, \texttt{age}=22, \texttt{education} = 3, \texttt{income}=80k)
  \end{align*}
  Then the set of grounded rules $C_{\bm x}$ is obtained from the
  rules~\eqref{eq:plaf:1}-\eqref{eq:plaf:3}.  For example,
  $\texttt{x\_cf.age $>=$ x.age}$ becomes $\texttt{age} \geq 22$.  The
  three grounded rules are:
  \begin{align}
    & \texttt{gender} = \texttt{female} \label{eq:plaf:1:grounded}  \\
    & \texttt{age} \geq 22 \label{eq:plaf:2:grounded}  \\
    & \texttt{education} > 3 \Rightarrow \texttt{age} > 26 \label{eq:plaf:3:grounded}
  \end{align}
  Every candidate $\bm x'$ must satisfy all three rules.
\end{example}

A naive strategy to generate candidates $\bm x'$ that satisfy
$C_{\bm x}$ is through rejection sampling: after each mutation and/or
crossover, we verify $C_{\bm x}$, and reject $\bm x'$ if it fails some
constraint in $C_{\bm x}$.  \geco\ improves over this naive approach
in two ways.  First, it computes for each group $G_i \in \Gamma$ a set
of values $DG_i$ that, in isolation, satisfy $C_{\bm x}$; this is
precomputed at the beginning of the algorithm by
\textbf{feasibleSpace}.  Second, once it generates candidates $\bm x'$
that differ in multiple feature groups $G_{i_1}, G_{i_2}, \ldots$ from
$\bm x$, then it enforces the constraint by possibly applying
additional mutation, until all constraints hold: this is done by the
function \textbf{actionCascade}.

The function $\textbf{feasibleSpace}(D, \Gamma, \texttt{C}_{\bm x})$
computes, for each group $G_i \in \Gamma$, the relation:
\begin{align*}
DG_i \defeq & \sigma_{C_{\bm x}^{(i)}}\left(\Pi_{G_i}(D)\right)
\end{align*}
where $C_{\bm x}^{(i)}$ consists of the conjunction of all rules in
$C_{\bm x}$ that refer only to features in the group $G_i$.  We call
the relation $DG_i$ the {\em sample space} for the group $G_i$.  Thus,
the selection operator $\sigma_{C_{\bm x}^{(i)}}$ rules out values in the sample
space that violate of some PLAF rule.

\begin{example}\label{ex:feature-space} Continuing
  Example~\ref{ex:plaf-grounded}, there are three groups,
  $G_1 = \set{\texttt{gender}}$,
  $G_2 = \set{\texttt{education},\texttt{income}}$,
  $G_3 = \set{\texttt{age}}$, and their sample spaces are computed as
  follows:
  \begin{align*}
    DG_1 \defeq & \sigma_{\texttt{gender} = \texttt{female}}(\Pi_{\texttt{gender}}(D))\\
    DG_2 \defeq & \Pi_{\texttt{education},\texttt{income}}(D))\\
    DG_3 \defeq & \sigma_{\texttt{age} \geq 22}(\Pi_{\texttt{age}}(D))
  \end{align*}
  Notice that we could only check the grounded
  rules~\eqref{eq:plaf:1:grounded} and~\eqref{eq:plaf:2:grounded}.
  The rule~\eqref{eq:plaf:3:grounded} refers to two different groups,
  and can only be checked by examining features from two groups; this
  is done by the function \textbf{actionCascade}.
\end{example}


The function
$\textbf{actionCascade}(\bm x', \Delta', \texttt{C}_{\bm x})$ is
called during mutation and crossover, and its role is to enforce the
grounded constraints $C_{\bm x}$ on a candidate $\bm x'$ before it is
added to the population.  Recall from Sec.~\ref{sec:constraints} that
the PLAF rules are acyclic.  \textbf{actionCascade} checks each rule,
in topological order of the acyclic graph: if the rule is violated,
then it changes the feature defined by that rule to a value that
satisfies the condition, and it adds the updated feature (or group) to
$\Delta'$.  The function returns the updated candidate $\bm x'$, which
now satisfies all rules $\texttt{C}_{\bm x}$, as well as its set of
changed features $\Delta'$.  For a simple illustration, referring to
Example~\ref{ex:feature-space}, when \geco\ considers a new candidate
$\bm x'$, it checks the rule~\eqref{eq:plaf:3:grounded}: if the rule
is violated, then it replaces $\texttt{age}$ with a new value from
$DG_3$, subject to the additional condition $\texttt{age} > 26$, and
adds $\texttt{age}$ to $\Delta'$.

\subsection{Selecting Fittest Candidates }
\label{sec:selection}

At each iteration, \geco's genetic algorithm extends the current population
(through mutation and crossover), then retains only the ``fittest'' $q$
candidates $\bm x'$ for the next generation using the
$\textbf{selectFittest}(\bm x, \pop, M, q)$ function. The function first
evaluates the fitness of each candidate $\bm x'$ in $\pop$; then sorts the
examples $\bm x'$ by their fitness score; and returns the top $q$ candidates.

We describe here how \geco computes the fitness of each candidate $\bm x'$. For
that, \geco takes into account two pieces of information: whether $\bm x'$ is
counterfactual, i.e. $M(\bm x')>0.5$, and the distance from $\bm x$ to $\bm x'$
which is given by the function $dist(\bm x, \bm x')$ defined in
Sec.~\ref{sec:problem} (see Eq.~\eqref{eq:dist}). It combines these two pieces
of information into a single numerical value, $score(\bm x')$ defined
as:
\begin{align}
  score(\bm x') =\! \begin{cases}
    dist(\bm x, \bm x') & \!\!\text{if } M(\bm x') > 0.5 \\
    \left(dist(\bm x, \bm x') + 1\right)\!+\!\left(1 - M(\bm x')\right) & \!\!\text{otherwise} \\
  \end{cases}
  \label{eq:score}
\end{align}
The rationale for this score is that we want every counterfactual
candidate to be better than every non-counterfactual candidate.  If
$M(\bm x') > 0.5$, then $\bm x'$ is counterfactual, in that case it
remains to minimize $dist(\bm x, \bm x')$.  But if
$M(\bm x') \leq 0.5$, then $\bm x'$ is not counterfactual, and we add
$1$ to $dist(\bm x, \bm x')$ ensuring that this term is larger than
any distance $dist(\bm x, \bm x'')$ for any counterfactual $\bm x''$
(because $dist \leq 1$, see Sec.~\ref{sec:problem}); we also add a
penalty $1-M(\bm x')$, to favor candidates $\bm x'$ that are closer to
becoming counterfactuals.

In summary, the function $\textbf{selectFittest}$ computes the fitness score
(Eq.~\eqref{eq:score}) for each candidate in the population, then sorts the
population in decreasing order by this score, and returns the $q$ fittest
candidates.  Notice that all counterfactuals examples $\bm x'$ will precede all
non-counterfactuals in the sorted order.

\begin{algorithm}[t]\centering
  \begin{tabular}{|p{0.95\columnwidth}|}\hline
    {\bf mutate} ($\pop, \Gamma, \texttt{DG}, \texttt{C}_{\bm x}, m$ )\\\hline\\[-0.8em]
    $\dcand  = \emptyset$ \\
    $\FOREACH\STAB (\bm x^*, \Delta^*) \in \pop\STAB\DO\STAB\{$\\
    $\TAB\FOREACH\STAB G \in \Gamma $ s.t. $G \notin \Delta^* \STAB\DO\STAB\{$\\
    $\TAB\TAB \bm x' = \bm x^*;\STAB S = \textbf{sample}(\texttt{DG}[G], m)$ \hfill{\small\texttt{//without replacement}} \\
    $\TAB\TAB\FOREACH\STAB v \in S \STAB\DO\STAB\{$ \\
    $\TAB\TAB\TAB \bm x'.G = v; \TAB \Delta' = \Delta^* \cup \set{G} $ \\
    $\TAB\TAB\TAB (\bm x', \Delta') = \textbf{actionCascade}(\bm x',\Delta', \texttt{C}_{\bm x})$ \\
    $\TAB\TAB\TAB\dcand \unioneq \set{(\bm x', \Delta')}$ \\[-0.3em]
    $\TAB\TAB\}$\\[-0.3em]
    $\TAB\}$\\[-0.3em]
    $\}$\\[-0.3em]
    $\RETURN\ \dcand$ \\\hline
  \end{tabular}
  \caption{Pseudo-code for \geco's mutate operator. }
\label{alg:mutation}
\end{algorithm}

\subsection{Mutation Operator}
\label{sec:mutation}

Algorithm~\ref{alg:mutation} provides the pseudo-code of the mutation
operator.  The operator
$\textbf{mutate}(\pop, \Gamma, \texttt{DG}, \texttt{C}_{\bm x}, m)$
takes as input the current population $\pop$, the list of feature
groups $\Gamma = \set{G_1, G_2, \ldots}$, their associated sample
spaces $\texttt{DG} = \set{DG_1, DG_2, \ldots}$, the grounded
constraints $\texttt{C}_{\bm x}$, and an integer $m > 0$.  The
function generates, for each candidate $\bm x'$ in the population, $m$
new mutations for each feature group $G_i$.  We briefly explain the
pseudo-code.  For each candidate $(\bm x^*, \Delta^*) \in \pop$, and
for each feature group $G \in \Gamma$ that has not been mutated yet
($G \not\in \Delta^*$), we sample $m$ values without replacement from
the sample space associated to $G$, and construct a new candidate
$\bm x'$ obtained by changing $G$ to $v$, for each value $v$ in the
sample.  As explained earlier (Sec.~\ref{sec:sample-space}), before we
insert $\bm x'$ in the population, we must enforce all constraints in
$C_{\bm x}$, and this is done by the function
$\textbf{actionCascade}$.

In \geco, the mutation operator also generates the initial
population. In this case, the current population $\pop$ contains only
the original instance $\bm x$ with $\Delta = \emptyset$. Thus, \geco
first explores candidates in the initial population that differ from the original instance $\bm x$ only in a
change for one group $G \in \Gamma$. This ensures that we prioritize
candidates that have few changes. Subsequent calls to the mutation
operator than change one additional feature group for each candidate
in the current population.

\begin{algorithm}[t]\centering
  \begin{tabular}{|p{0.95\columnwidth}|}\hline
    {\bf crossover} (\pop, $\texttt{C}_{\bm x}$ )\\\hline\\[-0.8em]
    $\dcand = \emptyset; \TAB\STAB \{\Delta_1, \ldots, \Delta_r\} = \{\Delta \mid (\bm x, \Delta) \in \pop \}$ \\
    $\FOREACH\STAB (\Delta_i, \Delta_j)  \in \{\Delta_1, \ldots, \Delta_r\} $ s.t. $ i < j \STAB\DO\STAB\{$ \\
    $\TAB \LET\ (\bm x_i, \bm \Delta_i) = \text{best instance in \pop with }\Delta = \Delta_i$\\
    $\TAB \LET\ (\bm x_j, \bm \Delta_j) = \text{best instance in \pop with }\Delta = \Delta_j$\\[0.3em]
    $\TAB$\texttt{//combine actions from $\bm x_i$ and $\bm x_j$}\\
    $\TAB\bm x' = \bm x_i; \TAB \bm \Delta' = \bm \Delta_i\cup \bm \Delta_j$\\
    $\TAB\FOREACH\STAB G \in \bm \Delta' \STAB\DO\STAB\{$ \\
    $\TAB\TAB\IF\STAB G \in \Delta_i \setminus \Delta_j \STAB\DO\STAB \bm x'.G = \bm x_i.G$ \\
    $\TAB\TAB\ELSE\IF\STAB G \in \Delta_j \setminus \Delta_i \STAB\DO\STAB \bm x'.G = \bm x_j.G$ \\
    $\TAB\TAB\ELSE\STAB \bm x'.G = \textbf{rand}(\texttt{Bool})\ ?\ \bm x_i.G : \bm x_j.G$ \\[-0.3em]
    $\TAB\}$\\
    $\TAB (\bm x', \Delta') = \textbf{actionCascade}(\bm x', \Delta', \texttt{C}_{\bm x})$ \\
    $\TAB\dcand \unioneq \set{(\bm x', \Delta')}$ \\[-0.3em]
    $\}$\\[-0.3em]
    $\RETURN\ \dcand$ \\\hline
  \end{tabular}
  \caption{Pseudo-code for  \geco's crossover operator.}
\label{alg:crossover}
\end{algorithm}

\subsection{Crossover Operator}
\label{sec:crossover}

The crossover operator generates new candidates by combining the
actions of two instances $\bm x_i, \bm x_j$ in $\pop$. For example,
consider candidates $\bm x_i$ and $\bm x_j$ that differ from $\bm x$
in the feature groups $\set{\texttt{age}}$, and
$\set{\texttt{education},\texttt{income}}$ respectively.  Then, the
crossover operator generates a new candidate $\bm x'$ that changes all
three features \texttt{age}, \texttt{education}, \texttt{income}, such
that that $\bm x'.\texttt{age} = \bm x_i.\texttt{age}$ and
$\bm x'.\set{\texttt{education},\texttt{income}} = \bm
x_j.\set{\texttt{education},\texttt{income}}$.  The pseudo-code of the
operator is given by Algorithm~\ref{alg:crossover}.

In a conventional genetic algorithm, the candidates for crossover are selected
at random. In \geco, however, we want to combine candidates that (1) change a
distinct set of features, and (2) are the best candidates amongst all candidates
in $\pop$ that change the same set of features. Recall that, for each candidate
$\bm x'$ in the population, we also store the set $\Delta'$ of feature groups
where $\bm x'$ differs from $\bm x$.  To achieve our customized crossover, we
first collect all distinct sets $\Delta$ in $\pop$. Then, we consider any
combination of two distinct sets $(\Delta_i, \Delta_j)$, and find the candidates
$\bm x_i$ and $\bm x_j$ which have the best fitness score (Eq.~\eqref{eq:score})
in the subset of $\pop$ for which $\Delta = \Delta_i$ and respectively $\Delta =
\Delta_j$.  Since $\pop$ is already sorted by the fitness score, the best
candidate with $\Delta = \Delta_i$ is also the first candidate in $\pop$ with
$\Delta = \Delta_i$. The operator then combines the candidates $\bm x_i$ and
$\bm x_j$ into a new candidate $\bm x'$, which inherits both the mutations from
$\bm x_i$ and $\bm x_j$; if a feature group $G$ was mutated in both $\bm x_i$
and $\bm x_j$, then we choose at random to assign to $\bm x'.G$ either $\bm
x_i.G$ or $\bm x_j.G$.  Finally, we check if $\bm x'$ requires any cascading
actions, and apply them in order to satisfy the constraints $C_{\bm x}$.  We add
$\bm x'$ and the corresponding feature set $\Delta'$ to the collection of new
candidates $\dcand$.

\subsection{Discussion}
\label{sec:implementation-discussion}

In this section, we discuss implementation details and potential tradeoffs that
we face for the performance optimization of \geco.

{\em Hyperparameters.} We use the following defaults for the
hyperparameters:
$q = 100, k=5, m_{\text{init}} = 20, m_{\text{mut}} = 5$.  The first
parameter means that each generation retains the fittest $q=100$
candidates.  The value $k=5$ means that the algorithm runs until the
top 5 candidates of the population are all counterfactuals (i.e.
$M(\bm x)>0.5$) {\em and} they remain in the top 5 during two
consecutive generations.  The initial population consists of $20$
candidates $\bm x'$ {\em per feature group}, and, during mutation, we
create 5 new candidates per feature group.  These defaults ensure that
selected candidates of the initial population will change at least
five distinct feature groups.

\ignore{
}

{\em Representation of $\Delta$.} For each candidate $(\bm x', \Delta') \in
\pop$, we represent the feature set $\Delta'$ as a compact bitset, which allows
for efficient bitwise operations.

{\em Sampling Operator.} In the mutation operator, we use weighted
sampling to sample actions from the sample space $DG_i$ associated to
the group $G_i$.  The weight is given by the frequency of each value
$v \in DG_i$ in the input dataset $D$. As a result, \geco is more
likely to sample those actions that frequently occur in the dataset
$D$, which helps to ensure that the generated counterfactual is
plausible.

{\em Number of Mutated Candidates.} By default the mutation operator mutates
every candidate in the current population, which ensures that (1) we explore a
large set of candidates, and (2) we sample many values from the feasible space
for each group. The downside is that this operation can take a long time, in
particular if we return many selected candidates ($q$ is large), and the
mutation operator can become a performance bottleneck. In this case, it is
possible to mutate only a selected number of candidates. We propose to group the
candidates by the set $\Delta$, and then to mutate only the top candidates in
each group (just like we only do crossover on the top candidate in each group).
This approach ensures that we still explore candidates with the $\Delta$ sets as
the default, but limits the total number of candidates that are generated.

{\em Large Sample Spaces.} If there is a very large sample space $DG_i$, then it
is possible that \geco does not sample the best action from this space. In this
case, we designed {\em selectiveMutate}, a variant of the mutate operator.
Whereas the normal mutation operator mutates candidates by changing feature
groups that have not been changed, selective mutation mutates feature groups
that were previously changed by sampling actions that decrease the distance
between the candidate and the original instance. This ensures that \geco is more
likely to identify the best action in large sample spaces.

\subsection{Comparison with CERTIFAI}
\label{sec:conv-ga}

CERTIFAI~\cite{certifai} also computes counterfactual explanations
using a genetic algorithm.  The main difference that distinguishes
CERTIFAI from \geco\ is that CERTIFAI assumes that the initial
population is a random sample of only ``good'' counterfactuals. Once
this initial population is computed, the goal of its genetic algorithm
is to find better counterfactuals, whose distance to the original
instance is smaller.  However, it is unclear how to compute the
initial population of counterfactuals, and the quality of the final
answer of CERTIFAI depends heavily on the choice of this initial
population.  CERTIFAI also does not emphasize the exploration of
counterfactuals that change only few features.  In contrast, \geco
starts from only the original instance $\bm x$, whose outcome
is ``bad'', and assumes that some ``good''
counterfactual is nearby, i.e. with few changed features.

Since CERTIFAI is not publicly available, we implemented our own
variant based on the description in the paper. Since it is not clear
how the initial population is computed, we consider all instances in
the database $D$ that are classified with the good outcome and satisfy
all feasibility and plausibility constraints. This is the most
efficient way to generate a sample of good counterfactuals, but it has
the downside is that the quality of the initial population, and,
hence, of the final explanation, varies widely with the instance
$\bm x$.  In fact, there are cases where $D$ contains no
counterfactual that satisfies the feasibility constraints w.r.t.
$\bm x$. In Section~\ref{sec:exp}, we show that \geco is able to
compute explanations that are closer to the original instance
significantly faster than CERTIFAI.

%% file: include/sec5-optimizations.tex
\newcommand{\drep}{\ensuremath{\Delta}-representation\xspace}

\section{Optimizations}
\label{sec:optim}

The main performance limitation in \geco are the repeated calls to
\textbf{selectFittest} and \textbf{mutate}. Between the two operations, \geco
repeatedly adds tens of thousands of candidates to the population, applies the
classifier $M$ on each of them, and then removes those that have a low fitness
score. In Section~\ref{sec:exp}, we show that selection and mutation account for
over 95\% of the overall runtime. In order to apply \geco in interactive
settings, it is thus important to optimize the performance of these two
operators. In this section, we present two optimizations which significantly
improve their performance.

To optimize {\bf mutate}, we present a loss-less, compressed data
representation, called \drep, for the candidate population that is generated
during the genetic algorithm. In our experiments, the \drep is up to $25\times$
more compact than an equivalent naive listing representation, which translates
to a performance improvement of up to 3.9$\times$ for mutate.

To optimize {\bf selectFittest}, we draw on techniques from the PL community, in
particular {\em partial evaluation}, to optimize the evaluation of the
classifier. The optimizations exploit the fact that we know the values for
subsets of the input features before the evaluation, which allows us to
translate the model into a specialized equivalent model that pre-evaluates the
static components. These optimizations improve the runtime for selectFittest by
up to $3.2\times$.

The $\Delta$-representation and partial evaluation complement each other, and
together decrease the end-to-end runtime of \geco by a factor of $5.2\times$.
Next, we provide more details for our optimizations.

\begin{figure}[t]
  \small
    \begin{tikzpicture}[scale=0.9]
      \matrix (df) at (0,0)
      [matrix of nodes, nodes in empty cells, ampersand replacement=\&]
      {
        Age  \& Edu. \& Occup. \& Income \\
        {\color{oxfordblue}28} \& HS \&  Service \& 10,000 \\
        {\color{oxfordblue}30} \& HS \&  Service \& 10,000 \\
        {\color{oxfordblue}32} \& HS \& Service \& 10,000 \\
        22 \& {\color{dgreen}PhD} \& Service \& 10,000 \\
        22 \& {\color{dgreen}BSc} \& Service \& 10,000 \\
        {\color{dred}25} \& HS \&  Service \& {\color{dred}15,000} \\
        {\color{dred}23} \& HS \&  Service \& {\color{dred}20,000} \\
        22 \& HS \& {\color{teal}Sales} \& 10,000 \\
        22 \& HS \& {\color{teal}Student} \& 10,000 \\
       };

       \draw ($(df-1-1.north west)$)-- ++(3.85,0);
       \draw ($(df-1-1.south west)+(0,0.05)$)-- ++(3.85,0);
       \draw (df-1-1.north west) -- ++(0,-4.6);
       \draw (df-1-1.north east) -- ++(0,-4.6);
       \draw (df-1-2.north east) -- ++(0,-4.6);
       \draw (df-1-3.north east) -- ++(0,-4.6);
       \draw (df-1-4.north east) -- ++(0,-4.6);
       \draw ($(df-1-1.north west)+(0,-4.6)$)-- ++(3.85,0);

      \matrix (dict) at (3.5,1.25)
      [matrix of nodes,
        nodes in empty cells,
        ampersand replacement=\&,
        column 1/.style={anchor=base west}]
      {
        $\Delta$ Sets \\
        \{Age\} \\
        \{Edu.\} \\
        \{Age, Income\} \\
        \{Occup.\}\\
       };

       \draw ($(dict-2-1.north west)$)-- ($(dict-5-1.south west)$);
       \draw ($(dict-2-1.north west)$)-- ++(2,0);
       \draw ($(dict-2-1.south west)$)-- ++(2,0);
       \draw ($(dict-3-1.south west)$)-- ++(2,0);
       \draw ($(dict-4-1.south west)$)-- ++(2,0);
       \draw ($(dict-5-1.south west)$)-- ++(2,0);
       \draw ($(dict-2-1.north west)+(2,0)$)-- ($(dict-5-1.south west)+(2,0)$);

      \matrix (age_df) at (6.8,1.18)
      [matrix of nodes,
        nodes in empty cells,
        ampersand replacement=\&]
      {
        Age  \\
        {\color{oxfordblue}28} \\
        {\color{oxfordblue}30} \\
        {\color{oxfordblue}32} \\
      };

      \draw ($(age_df-1-1.north west)$) -- ++(0,-1.8);
      \draw ($(age_df-1-1.north east)$) -- ++(0,-1.8);
      \draw ($(age_df-1-1.north west)$) -- ($(age_df-1-1.north east)$);
      \draw ($(age_df-1-1.south west)$) -- ($(age_df-1-1.south east)$);
      \draw ($(age_df-1-1.north west)+(0,-1.8)$) -- ($(age_df-1-1.north east)+(0,-1.8)$);

      \matrix (edu_df) at (5.7,0.4)
      [matrix of nodes,
        nodes in empty cells,
        ampersand replacement=\&]
      {
        Edu.  \\
        {\color{dgreen}PhD} \\
        {\color{dgreen}BSc} \\
      };

      \draw ($(edu_df-1-1.north west)$) -- ++(0,-1.4);
      \draw ($(edu_df-1-1.north east)$) -- ++(0,-1.4);
      \draw ($(edu_df-1-1.north west)$) -- ($(edu_df-1-1.north east)$);
      \draw ($(edu_df-1-1.south west)$) -- ($(edu_df-1-1.south east)$);
      \draw ($(edu_df-1-1.north west)+(0,-1.4)$) -- ($(edu_df-1-1.north east)+(0,-1.4)$);

      \matrix (inc_df) at (5.5,-1.4)
      [matrix of nodes,
        nodes in empty cells,
        ampersand replacement=\&]
      {
        Age \& Income \\
        {\color{dred}25} \& {\color{dred}15,000} \\
        {\color{dred}23} \& {\color{dred}20,000} \\
      };

      \draw ($(inc_df-1-1.north west)$) -- ++(0,-1.4);
      \draw ($(inc_df-1-1.north east)$) -- ++(0,-1.4);
      \draw ($(inc_df-1-1.north west)$) -- ++(1.9,0);
      \draw ($(inc_df-1-1.south west)$) -- ++(1.9,0);
      \draw ($(inc_df-1-1.north west)+(1.9,0)$) -- ++(0,-1.4);
      \draw ($(inc_df-1-1.north west)+(0,-1.4)$) -- ++(1.9,0);

      \matrix (occ_df) at (3.5,-1.2)
      [matrix of nodes,
        nodes in empty cells,
        ampersand replacement=\&]
      {
        Occup.\\
        {\color{teal}Sales} \\
        {\color{teal}Student} \\
      };

      \draw ($(occ_df-1-1.north west)$) -- ++(0,-1.4);
      \draw ($(occ_df-1-1.north east)$) -- ++(0,-1.4);
      \draw ($(occ_df-1-1.north west)$) -- ($(occ_df-1-1.north east)$);
      \draw ($(occ_df-1-1.south west)$) -- ($(occ_df-1-1.south east)$);
      \draw ($(occ_df-1-1.north west)+(0,-1.4)$) -- ($(occ_df-1-1.north east)+(0,-1.4)$);

      \draw[->] ($(dict-2-1.west)+(2,0)$) -- (age_df-1-1.west);
      \draw[->] ($(dict-3-1.west)+(2,0)$) -| (edu_df-1-1.north);
      \draw[->] ($(dict-4-1.west)+(2,0)$) -| ($(inc_df-1-1.north)+(-0.1,0)$);
      \draw[->] ($(dict-5-1.south)+(.35,0)$) -- (occ_df-1-1.north);

    \end{tikzpicture}
    \vspace*{-1em}

    \caption{Example candidate population for instance $\bm x =
      (\text{22,HS,Service,10000})$ using the naive listing representation
      (left) and the equivalent $\Delta$-representation (right).}
    \label{fig:deltarep}
\end{figure}
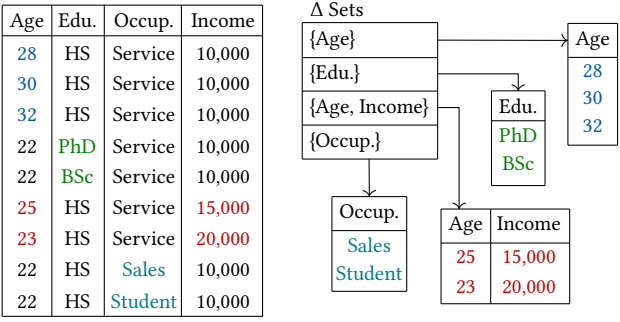

\subsection{$\Delta$-Representation}
\label{sec:drep}

The naive representation for the candidate population of the genetic algorithm
is a listing of the full feature vectors $\bm x'$ for all candidates $(\bm x',
\Delta')$. This representation is highly redundant, because most values in $\bm
x'$ are equal to the original instance $\bm x$; only the features in $\Delta'$
are different. The {\em $\Delta$-representation} can represent each the
candidate  $(\bm x', \Delta')$ compactly by storing only the features in
$\Delta'$. This is achieve by grouping  the candidate population by the set of
features $\Delta'$, and then representing the entire subpopulation in a single
relation $R_{\Delta'}$ whose attributes are only $\Delta'$.

\begin{example}
  Figure~\ref{fig:deltarep} presents a candidate population for the instance
  $\bm x = (\texttt{Age=}22, \texttt{Edu=}\text{HS},
  \texttt{Occup=}\text{Service},\texttt{Income=}10000)$ using (left) the naive
  listing representation and (right) the equivalent $\Delta$-representation. For
  simplicity, we highlight the changed values in the listing representation,
  instead of enumerating all $\Delta$ sets,

  Most values stored in the listing representation are values from $\bm x$. In
  contrast, the $\Delta$-representation only represents the values that are
  different from $\bm x$. For instance, the first three candidates, which change
  only \texttt{Age}, are represented in a relation with attributes \texttt{Age}
  only, and without repeating the values for \texttt{Edu}, \texttt{Occup},
  \texttt{Income}.
\end{example}

In our implementation, we represent the $\Delta$ sets as bitsets, and the
$\Delta$-representation is a hashmap of that maps the distinct $\Delta$ sets to
the corresponding relation $R_\Delta$, which is represented as a DataFrame. We
provide wrapper functions so that we can apply standard DataFrame operations
directly on the $\Delta$-representation.

The $\Delta$-representation has a significantly lower memory footprint, which
can lead to significant performance improvements over the naive representation,
because it is more efficient to add candidates to the smaller relations, and it
simplifies garbage collection.

There is, however, a potential performance tradeoff for the selection operator,
because the classifier $M$ typically assumes as input the full feature vector
$\bm x'$. In this case, we copy the values in the $\Delta$-representation to a
full instantiation of $\bm x'$. This transformation can be expensive, but, in
our experiments, it does not outweigh the speedup for mutation. In the next
section, we show how we can use partial evaluation to avoid the construction of
the full $\bm x'$.

\subsection{Partial Evaluation for Classifiers}

We show how to adapt PL techniques, in particular code specialization via
partial evaluation, to optimize the evaluation of a given classifier $M$, and
thus speedup the performance of {\bf selectFittest}.

Consider a program $P : X \times Y \to O$ which maps two inputs $(X, Y)$ into
output $O$. Assume that we know $Y = y$ at compile time. Partial evaluation
takes program $P$ and input $Y = y$ and generates a more efficient program
$P'_{\langle y \rangle} X \to O$, which precomputes the static components.
Partial evaluation guarantees that $P'_{\langle y \rangle}(x)  = P(x,y)$ for all
$x \in \text{dom}(X)$. See~\cite{jones1993partial} for more details on partial
evaluation.

We next overview how we use partial evaluation in \geco. Consider a classifier
$M$ with features $F$. During the evaluation of candidate $(\bm x', \Delta')$,
we know that the values for all features $F \setminus \Delta'$ are constants
taken from the original instance $\bm x$. Thus, we can partially evaluate the
classifier $M$ to a simplified classifier $M_{\Delta'}$ that precomputes the
static components related to features $F \setminus \Delta'$. Once $M_{\Delta'}$
is generated, we cache the model so that we can apply it for all candidates in
the population that change the same feature set $\Delta'$. Note that by using
partial evaluation, \geco no longer explains a black-box, since it requires
access to the code of the classifier.

\begin{example}
  Consider a decision tree classifier $M$. For an instance $(\bm x', \Delta')$,
  $M(\bm x')$ typically evaluates the decisions for all features along a root to
  leaf path. Since we know the values for the features $F \setminus \Delta'$, we
  can precompute and fold all nodes in the tree that involve the features $F
  \setminus \Delta'$. If $\Delta'$ is small, then partial evaluation can
  generate a very simple tree. For instance, if $\Delta' = \{\texttt{Age} \}$
  then $M_{\Delta'}$ only needs to evaluate decisions of the form $\textit{age} >
  30$ or $30 < \textit{age} < 60$ to classify the input.
\end{example}

In addition to optimizing the evaluation of $M$, a partially evaluated
classifier $M_{\Delta}$ can be directly evaluated over the partial relation
$R_\Delta$ in the $\Delta$-representation, and thus we mitigate the overhead
resulting from the need to construct the full entity for the evaluation.

Partial evaluation has been studied and applied in various domains, e.g., in
databases, it has been used to optimize query evaluation (see
e.g.,~\cite{Shaikhha:2018,10.1145/3183713.3196893}). We are, however, not aware of a
general-purpose partial evaluator that be applied in \geco to optimize arbitrary
classifiers. Thus, we implemented our own partial evaluator for two model
classes: (1) tree-based models, which includes decision trees, random forests,
and gradient boosted trees, and (2) neural networks and multi-layered
perceptrons.

In the following, we briefly introduce the partial evaluation we use for tree-based
models and neural networks.

\mysubsubsec{Tree-based models}
We optimize the evaluation of tree-based models in two steps. First, we use
existing techniques to turn the model into a more optimized representation for
evaluation. Then, we apply partial evaluation on the optimized representation.

Tree-based models face performance bottlenecks during evaluation because, by
nature of their representation, they are prone to cache misses and branch
misprediction. For this reason, the ML systems community has studied how
tree-based models can be represented so that they can be evaluated without
random lookups in memory and repeated if-statements (see
e.g.,~\cite{asadi2013runtime, lucchese2015quickscorer, ye2018rapidscorer}). In
\geco, we use the representation proposed by the QuickScorer
algorithm~\cite{lucchese2015quickscorer}, which we briefly overview next.

Instead of evaluating a decision tree $T$ following a root to leaf path,
QuickScorer evaluates {\em all} decision nodes in $T$ and keeps track of which
leaves cannot be reached whenever a decision fails. Once all decisions are
evaluated, the prediction is guaranteed to be given by the first leaf that can
be reached.  All operations in QuickScorer use efficient, cache-conscious
bitwise operations, and avoid branch mispredictions. This makes the QuickScorer
very efficient, even if it evaluates many more decisions than the naive
evaluation.

For partial evaluation, we exploit the fact that the first phase in QuickScorer
computes the decision nodes for each feature independently of all other
features. Thus, given a set of fixed feature values, we can precompute all corresponding
decisions, and significantly number of decision that are evaluated at runtime.

\mysubsubsec{Neural Networks and MLPs}
We consider neural networks and multi-layered perceptrons for structured,
tabular data (as opposed to images or text). In this setting, each hidden node
$N$ in the first layer of the network is typically a linear
model~\cite{arik2019tabnet,mahajan2019causalcf}. Given an input vector $\bm x$,
parameter vector $\bm w$, and bias term $b$, the node $N$ thus computes: $y =
\sigma( \bm x^\top \bm w + b)$, where $\sigma$ is an activation function. If we
know that some values in $\bm x$ are static, then we can apply partial
evaluation for $N$ by precomputing the product between $\bm x$ and $\bm w$ for
all static components and adding the partial product to the bias term $b$.
During evaluation, we then only need to compute the product between $\bm x $ and
$\bm w$ for the non-static components.

The impact of partial evalaution depends on the network structure.
When applied to structured data, the model typically consists of fully-connected
layers, in which case we can only partially evaluate the first layer of the
network. We can apply partial evaluation to subsequent layers, only if the
layers are not fully-connected.

\ignore{
}

%% file: include/sec6-experiments.tex
\begin{table}[tb]
  \caption{Key characteristics for each considered dataset.}
  \begin{tabular}{|l|c|c|c|c|} \hline
                               & Credit & Adult & Allstate & Yelp   \\\hline\hline
    Data Points                & 30K    & 45K   & 13.2M    & 22.4M  \\\hline
    Variables                  & 14     & 12    & 29       & 34     \\\hline
    Features (one-hot enc.)    & 14     & 42    & 548      & 764    \\\hline
    \hline
    Feature Groups             & 14     & 11    & 29       & 34     \\\hline
    Constraints / Implications & 7 / 2  & 7 / 1 & 0 / 0    & 18 / 2 \\\hline
  \end{tabular}
  \label{tbl:datasets}
\end{table}

\section{Experiments}
\label{sec:exp}

We present the results for our experimental evaluation of \geco on four real
datasets. We conduct the following experiments:

\begin{enumerate}
  \item We investigate whether \geco is able to compute counterfactual
        explanations for one end-to-end example, and compare the explanation
        with five existing systems.
  \item We benchmark all considered systems on 5,000 instances, and investigate
        the tradeoff between the quality of the explanations and the runtime for
        each system.
  \item We conduct microbenchmarks for \geco. In particular, we breakdown the
        runtime into individual components and investigate the impact of the
        optimizations from Section~\ref{sec:optim}. We also evaluate GeCo's
        ability to find the optimal explanation using synthetic classifiers.
\end{enumerate}


\subsection{Experimental Setup}
\label{sec:exp-setup}

In this section, we present the considered datasets and systems, as well as the
setup used for all our experiments.

\mysubsubsec{Datasets}
We consider four real datasets: (1) \emph{Credit}~\cite{credit-data} is used
predict customer's default on credit card payments in Taiwan; (2)
\emph{Adult}~\cite{adult-data} is used to predict whether the income of adults
exceeds \$50K/year using US census data from 1994; (3) \emph{Allstate} is a
Kaggle dataset for the Allstate Claim Prediction Challenge~\cite{allstate-data},
used to predict insurance claims based on the characteristics of the insured’s
vehicle; (4) \emph{Yelp} is based on the public Yelp Dataset
Challenge~\cite{yelp-data} and is used to predict review ratings that users give
to businesses.

Table~\ref{tbl:datasets} presents key statistics for each dataset. Credit and
Adult are from the UCI repository~\cite{Dua:2019} and commonly used to evaluate
explanations (e.g.,~\cite{mace,mothilal2020diverse,ustun:recourse}). For all
datasets, we one-hot encode the categorical variables. For the evaluation with
existing systems, we further apply the same preprocessing that was proposed by
the existing system, in order to ensure that our evaluation is fair.

For all datasets, we group the features derived from one-hot encoding in one
feature group. In addition, we encode various PLAF constraints with and without
implications (cf. Table~\ref{tbl:datasets}). For instance, we enforce that
\texttt{Age} and \texttt{Education} can only increase, and
\texttt{MaritalStatus}, \texttt{Gender}, and \texttt{NativeCountry} cannot
change. An example of a constraint with implications is given by Eq.
\eqref{eq:plaf:3}. We present a detailed description of all considered PLAF
constraints in Appendix~\ref{appendix:constraints}. Since
the existing systems do not support constraints with implications, we do not
enforce these constraints in the experiments in Sec.~\ref{sec:exp:tradeoff}.


\mysubsubsec{Considered Systems}
We benchmark \geco against five existing systems. (1) \emph{MACE}~\cite{mace}
solves for counterfactuals with multiple runs of an SMT solver. (2)
\emph{DiCE}~\cite{mahajan2019causalcf} generates counterfactual explanations
with a variational auto-encoder. (3) \emph{WIT} is our implementation of the
counterfactual reasoning approach in Google's What-if Tool~\cite{wexler2019wit}.
WIT looks up the closest counterfactual that satisfies the PLAF constraints in
database $D$. We implemented our own version, because the What-if Tool does not
support feasibility constraints. (4) \emph{CERT} is our implementation of the
genetic algorithm that is used in CERTIFAI~\cite{certifai} (see
Sec.~\ref{sec:conv-ga} for details). We reimplemented the algorithm because
CERTIFAI is not publicly available. (5) {\em SimCF} is our adaptation of the
SimBA~\cite{guo2019simba} algorithm for adversarial examples to the problem of
finding counterfactual explanations. The algorithm randomly selects one feature
group, samples five feasible values for this group, and greedily applies the
change that returns the best score. This process is repeated until the
classifier returns the desired outcome. Since SimCF randomly changes one feature
at a time, the explanations may not be consistent across runs.

\mysubsubsec{Evaluation Metrics} We use the following three metrics to evaluate
the quality of the explanation: (1) The {\em consistency} of the explanations,
i.e., does the classifier return the good outcome for the counterfactual $\bm
  x_{cf}$; (2) The {\em distance} between $\bm x_{cf}$ and the original instance
$\bm x$; (3) The {\em number of features changed} in $\bm x_{cf}$.

For the comparison with existing systems, we use the $\ell_1$ norm to aggregate
the distances for each feature (i.e., $\beta = 1$, $\alpha = \gamma = 0$ in
Eq.~\eqref{eq:dist}), because MACE and DiCE do not support combining norms. We
examine other choices of these hyperparameters in
Appendix~\ref{appendix:distance}. To compare runtimes,
we report the average wall-clock time it takes to explain a single instance.

\geco and CERT return multiple counterfactuals for each instance. We only
consider the best counterfactual in this evaluation.

\begin{table}[t]
  \small
  \caption{Examples of counterfactual explanations by \geco, MACE, DiCE, and
    SimCF for one instance in Adult (presenting selected features). MACE and
    DiCE use different models; we show \geco's explanation for each model. The
    neural network does not use CapitalGain and CapitalLoss.}
  \vspace*{-1.5em}

  \setlength\tabcolsep{4.3pt}
  \hspace*{-1.6em}

  \begin{tabular}{r|r|c|c|r|r|r|c|l}
    \multicolumn{1}{c}{}         & \rot{Age}                              & \rot{Education}           & \rot{Occupation}          & \rot{CapitalGain}       & \rot{CapitalLoss}       & \rot{Hours/week}     & \rot{Gender}        & \rot{Prediction}         \\
    \multicolumn{9}{c}{}                                                                                                                                                                                                                                      \\[-0.9em]\cline{2-8}
    \multicolumn{1}{c|}{$\bm x$} & 49                                     & School                    & Service                   & 0                       & 0                       & 16                   & F                   & {\color{dred}Bad}        \\\cline{2-8}
    \multicolumn{9}{c}{}                                                                                                                                                                                                                                      \\[-0.9em]
    \multicolumn{1}{c}{}         & \multicolumn{7}{c}{\bf Decision Tree}  &                                                                                                                                                                                   \\\cline{2-8} 
    \multicolumn{1}{c|}{\geco}   & 49                                     & School                    & Service                   & {\color{dred}\bf 4,787} & 0                       & 16                   & F                   & {\color{dgreen}\bf Good} \\\cline{2-8}
    \multicolumn{9}{c}{}                                                                                                                                                                                                                                      \\[-0.8em]\cline{2-8}
    \multicolumn{1}{c|}{MACE}    & 49                                     & School                    & Service                   & {\color{dred}\bf 4,826} & {\color{dred}\bf 20}    & 16                   & F                   & {\color{dgreen}\bf Good} \\\cline{2-8}
    \multicolumn{9}{c}{}                                                                                                                                                                                                                                      \\[-0.8em]\cline{2-8}
    \multicolumn{1}{c|}{SimCF}   & 49                                     & School                    & Service                   & {\color{dred}\bf 7,688} & {\color{dred}\bf 1,380} & 16                   & F                   & {\color{dgreen}\bf Good} \\\cline{2-8}
    \multicolumn{9}{c}{}                                                                                                                                                                                                                                      \\[-0.9em]
    \multicolumn{1}{c}{}         & \multicolumn{7}{c}{\bf Neural Network} &                                                                                                                                                                                   \\\cline{2-8} 
    \multicolumn{1}{c|}{\geco}   & {\color{dred}\bf 53}                   & {\color{dred}\bf Masters} & Service                   & --                      & --                      & 16                   & F                   & {\color{dgreen}\bf Good} \\\cline{2-8}
    \multicolumn{9}{c}{}                                                                                                                                                                                                                                      \\[-0.8em]\cline{2-8}
    \multicolumn{1}{c|}{DiCE}    & {\color{dred}\bf 54}                   & {\color{dred}\bf PhD}     & {\color{dred}\bf BlueCol} & --                      & --                      & {\color{dred}\bf 40} & {\color{dred}\bf M} & {\color{dgreen}\bf Good} \\\cline{2-8}
  \end{tabular}
  \label{tbl:explanatinon-examples}
\end{table}

\begin{figure*}[t]
  \centering

  {\bf Credit Dataset} \hspace{6.5cm}  {\bf Adult Dataset}

  \includegraphics[width=0.245\linewidth]{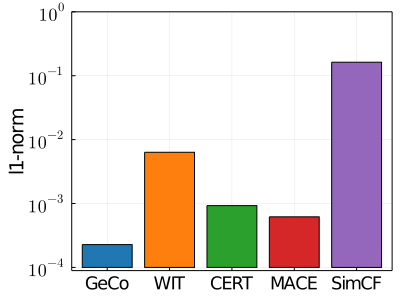}
  \includegraphics[width=0.245\linewidth]{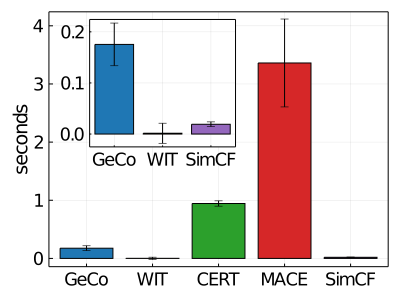}
  \includegraphics[width=0.245\linewidth]{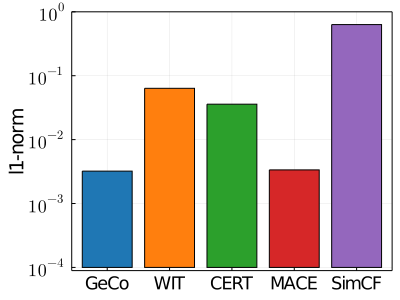}
  \includegraphics[width=0.245\linewidth]{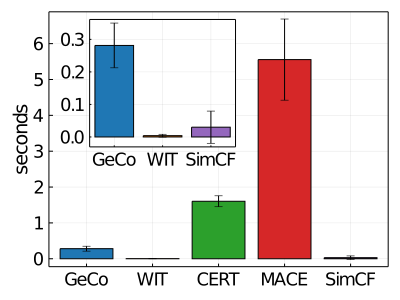}

  \vspace{-1em}

  \caption{Comparison of the average distance ($\ell_1$ norm, $\log_{10}$ scale),
    and average runtime for the explanations by \geco, WIT, CERT, MACE, and SimCF for
    5000 instances on the Credit and Adult datasets. Error bars represent one
    standard deviation. }
  \label{fig:exp1-dist}
\end{figure*}

\mysubsubsec{Classifiers}
We benchmark the systems on tree-based models and multi-layered perceptrons
(MLP).

  {\em Comparison with Existing Systems.} For the comparison of with existing
systems we use the classifiers proposed by MACE and DiCE for all systems to
ensure a fair comparison. For tree-based models, we attempted to compute
explanations for random forest classifiers, but MACE took on average 30 minutes
to compute a {\emph single explanation}, which made it infeasible to compute
explanations for many instances. Thus, we consider a single decision tree for
this evaluation (computed in scikit-learn, default parameters). DiCE does not
support decision trees, since it requires a differentiable classifier. For the
neural net, we use the classifier proposed by DiCE, which is a two-layered neural
network with 20 (fully connected) hidden units and ReLU activation.

  {\em Microbenchmarks.} In the micro benchmarks, we consider a random forest
with 500 trees and maximum depth of 10 from the Julia MLJ
library~\cite{blaom2020mlj}, and the MLPClassifier from the scikit-learn
library~\cite{scikit2011}. We learn two MLP models, a small variant
with one hidden layer (100 nodes, the default setting) and a larger variant with
two hidden layers (100 nodes each), which was the best network structure we
found in a comparison of 10 different structures.

\mysubsubsec{Setup}
We implemented \geco, WIT, CERT, and SimCF in Julia 1.5.2. All experiments are run on
an Intel Xeon CPU E7-4890/2.80GHz/ 64bit with 108GB RAM, Linux 4.16.0, and Ubuntu
16.04.

We use the default hyperparameters for \geco (c.f.
Sec.~\ref{sec:implementation-discussion}) and MACE ($\epsilon = 10^{-3}$). CERT
runs for 300 generations, as in the original CERTIFAI implementation. In \geco,
we precompute the active domain of each feature group, which is invariant for
all explanations.

\subsection{End-to-end Example}
\label{sec:exp:end-to-end}

We consider one specific instance in the Adult dataset that is classified as
``bad'' (\texttt{Income} <\$50K) and illustrate the differences between the
the explanations for each considered system.

Table~\ref{tbl:explanatinon-examples} presents the instance $\bm x$ and the
counterfactuals returned by \geco, MACE, SimCF, and DiCE. WIT and CERT fail to
return an explanation, because the Adult dataset has no instance with income
>\$50K that also satisfies all PLAF constraints for $\bm x$. MACE and SimCF
compute the explanation over a decision tree, and DiCE considers a neural
network. We present \geco's explanation for each model, and argue that they are
better than the explanations by MACE, SimCF, and DiCE.

For the decision tree, \geco is able to find a counterfactual that changes only
Capital Gains. In contrast, the explanations by MACE and SimCF require a change
in Capital Gains and Capital Loss. Remarkably, their changes in Capital Gains
are larger than the one required by \geco. The neural network does not use the
features Capital Gains and Capital Loss. For this model, \geco proposes an
increase in education, which in turn requires an increase in age according to
our PLAF constraints. If this change is deemed infeasible, we can update the
PLAF constraints and ask \geco to generate a new counterfactual. In contrast,
DiCE changes the values of eight features in total, which is neither feasible
nor plausible.

\ignore{
}

\ignore{





}

\subsection{Quality and Runtime Tradeoff}
\label{sec:exp:tradeoff}

In this section, we investigate the tradeoff between the quality and the runtime
of the explanations for all considered systems on Credit and Adult. As explained
in Sec.~\ref{sec:exp-setup}, we evaluate the systems using a single decision
tree and a neural network.

\mysubsubsec{Takeaways for Evaluation with Decision Trees}
Figures~\ref{fig:exp1-dist} shows the results of our evaluation with decision
trees on 5,000 instances from Credit and Adult for which the classifier returns
the negative outcome. We present the average distance and runtime for each
considered system and dataset.

\geco and MACE are always able to find a feasible and plausible explanation. WIT
and CERT, however, fail to find an explanation in 2.1\% of the cases for Adult.
This is because the two techniques are restricted by the database $D$, which may
not contain an instance that is classified as good and represents feasible and
plausible actions. SimCF fails to find explanations in 1.5\% and
  2.4\% of the cases for Adult and respectively Credit.

\geco's explanations are on average the closest to the original instance. This
can be explained by the fact that \geco is able to find these explanations by
changing significantly fewer features. For the Credit dataset, for instance,
\geco can find explanations with 1.27 changes on average, while MACE (the best
competitor) changes on average 3.39 features. WIT, CERT, and SimCF change on average
3.97, 3.15, and respectively 2.98 features. We provide further
  details on the number of features changed by each system in
  Appendix~\ref{appendix:numfeats}.

\geco is consistently able to compute each explanation in less than 300ms on
average. On average, \geco is $20\times$ faster than MACE. This
performance is only matched by WIT and SimCF, which do not return explanations
with the same quality as \geco.

Like \geco, CERT uses a genetic algorithm, but it takes $5.6\times$ longer to
compute explanations that do not have the same quality as \geco's. Thus, \geco's
custom genetic algorithm, which is designed to explore counterfactuals with few
changes, is very effective.

\begin{figure}
  \centering

  \includegraphics[width=0.45\columnwidth]{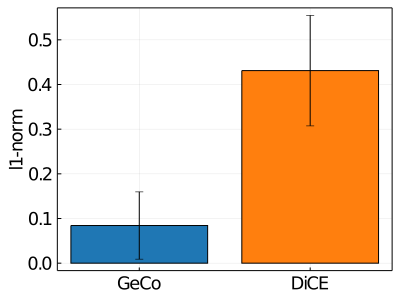}
  \hspace{0.3em}
  \includegraphics[width=0.45\columnwidth]{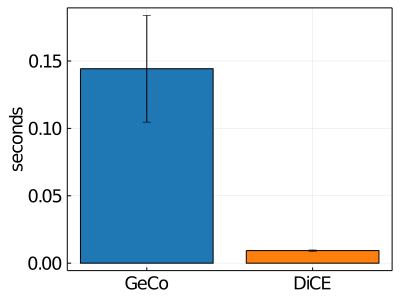}
  \vspace{-1em}

  \caption{Comparison of the average distance and runtime of \geco and DiCE
  for a neural network on 5000 Adult instances. Error bars represent one
  standard deviation.}
  \label{fig:exp1-dice}
\end{figure}

\mysubsubsec{Takeaways for Evaluation with Neural Net}
Figure~\ref{fig:exp1-dice} presents the key results for our evaluation on the
MLP classifier. We only show the comparison of \geco and DiCE, because the
comparison with WIT, SimCF, and CERT is similar to the one for decision
trees. MACE requires an extensive conversion of the classifier into a logical
formula, which is not supported for the considered model.

Since DiCE computes the counterfactual explanation in one pass over a
variational auto-encoder, it is able to compute the explanations very
efficiently. In our experiments, DiCE was able to find an explanation on average
$15.5\times$ faster than \geco. The counterfactuals that DiCE generates,
however, have poor quality. Whereas \geco is again able to find explanations
with only few required changes, DiCE changes on average 5.3 features. As a
result, \geco's explanation is on average $4.4\times$ closer to the original
instance. Therefore, we consider \geco much more suitable for real-world
applications.

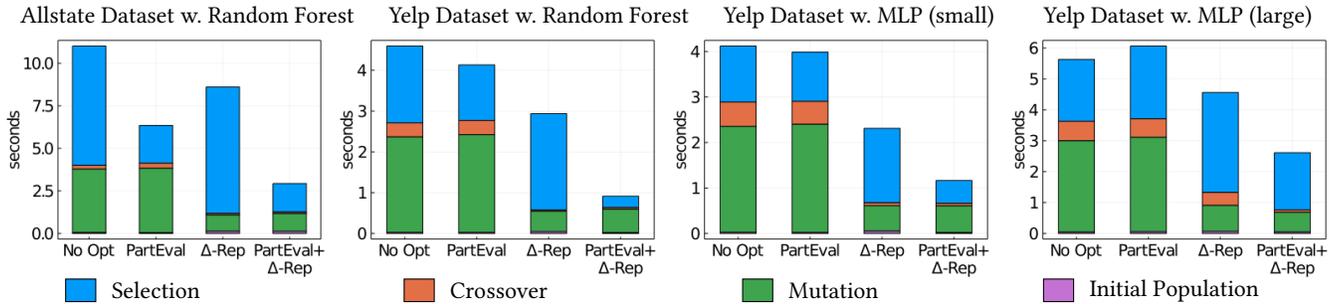
\begin{figure*}[t]

  Allstate Dataset w. Random Forest \hspace{.3cm} Yelp Dataset w. Random Forest \hspace{.3cm} Yelp Dataset w. MLP (small) \hspace{.5cm} Yelp Dataset w. MLP (large)

  \includegraphics[width=0.249\linewidth]
  {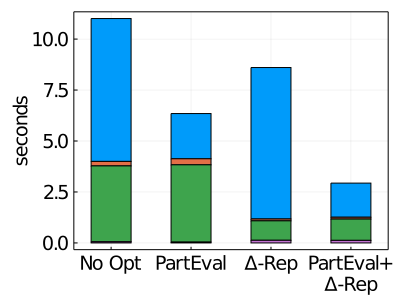}\hfill
  \includegraphics[width=0.249\linewidth]
  {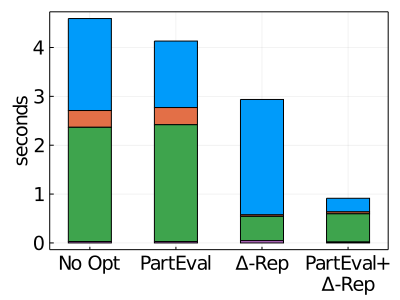}\hfill
  \includegraphics[width=0.249\linewidth]
  {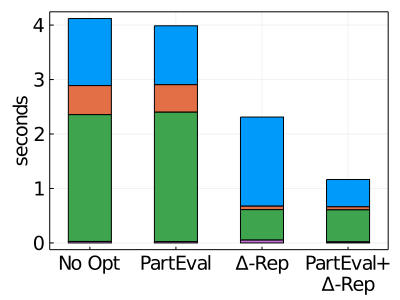}
  \includegraphics[width=0.249\linewidth]
  {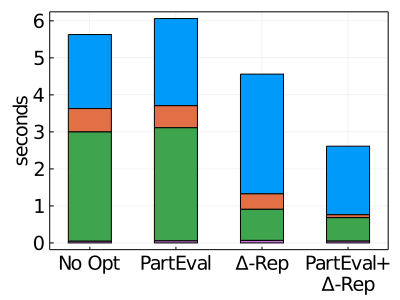}

  \vspace*{-.5em}

  \begin{tikzpicture}
    \coordinate (select) at (0,0);
    \draw[fill=plot1] ($(select)+(-.3,-.15)$) rectangle ($(select)+(.1,.15)$);
    \node[anchor=west,black] at ($(select)+(.2,0)$) {Selection};

    \coordinate (cross) at (4.5,0);
    \draw[fill=plot2] ($(cross)+(-.3,-.15)$) rectangle ($(cross)+(.1,.15)$);
    \node[anchor=west,black] at ($(cross)+(.2,0)$) {Crossover};

    \coordinate (mut) at (9,0);
    \draw[fill=plot3] ($(mut)+(-.3,-.15)$) rectangle ($(mut)+(.1,.15)$);
    \node[anchor=west,black] at ($(mut)+(.2,0)$) {Mutation};

    \coordinate (init) at (13,0);
    \draw[fill=plot4] ($(init)+(-.3,-.15)$) rectangle ($(init)+(.1,.15)$);
    \node[anchor=west,black] at ($(init)+(.2,0)$) {Initial Population};
  \end{tikzpicture}

  \vspace*{-1em}

  \Description[]{}

  \caption{Breakdown of \geco's runtime into the main operators on Allstate and
    Yelp using random forest and MLP classifiers. We present the runtimes (1)
    without the $\Delta$-representation and partial evaluation, (2) with
    partial evaluation, (3) with the $\Delta$-representation, and (4) with both
    optimizations. The runtime is averaged over 1,000 instances.}
  \label{fig:exp2-breakdown}
\end{figure*}

\begin{table}[t]
  \caption{Microbenchmarks results for tree-based models.}
  \begin{tabular}{|l|r|r|r|r|} \hline
                          & Credit      & Adult       & Allstate     & Yelp         \\\hline\hline
    Generations           & 4.22        & 4.66        & 5.26         & 3.24         \\\hline
    Explored Candidates   & 18.1K       & 15.3K       & 62.8K        & 42.9K        \\\hline\hline
    Size of Naive Rep.    & 62.23K      & 117.41K     & 8.03M        & 7.84M       \\\hline
    Size of $\Delta$-Rep. & 18.64K      & 21.88K      & 412K         & 131K         \\\hline
    Compression           & 3.3$\times$ & 5.4$\times$ & 19.5$\times$ & 60.1$\times$ \\\hline
  \end{tabular}
  \label{tbl:microbench}
\end{table}

\subsection{Microbenchmarks}
\label{sec:microbenchmarks}

In this section, we present the results for the microbenchmarks.

\mysubsubsec{Breakdown of \geco's Components}
First, we analyze the runtime of each operator presented in
Sec.~\ref{sec:algorithm}, as well as the impact of the $\Delta$-representation
and partial evaluation (Sec.~\ref{sec:optim}) on two tree-based models over
Allstate and Yelp, as well as the small and large multi-layered perceptrons
(MLP) over Yelp. We run \geco for 5 generations on 1,000 instances that have
been classified as bad.

Figure~\ref{fig:exp2-breakdown} presents the results for this benchmark. Initial
population captures the time it takes to compute the feasible space, and to
generate and select the fittest candidates for the initial population. The times
for selection, crossover, and mutation are accumulated over all generations. For
each scenario, we first present the runtime for: (1) without the
$\Delta$-representation and partial evaluation enabled, (2) with each
optimization individually, and (3) with both.

The results show that the selection and mutation operators are the most time
consuming operations of the genetic algorithm. This is not surprising since they
operate on tens of thousands of candidates, whereas crossover combines  only a
few selected candidates.

Partial evaluation of the classifier is effective for the random forest model.
For Allstate, it decreases the runtime of the selection operator by up to
3.2$\times$, which translates into an overall speedup of $1.7\times$. For MLPs,
the optimization is less effective, because we can only partially evaluate the
first layer. In fact, the overhead of partial evaluation results in a slowdown
for the larger variant.

The $\Delta$-representation decreases the runtime of the mutation operator by
3.9$\times$ for Allstate and $4.7\times$ for Yelp (random forest). This speedup
is due to the compression achieved by the $\Delta$-representation (see below).
If the classifier is not partially evaluated, then there is a tradeoff in the
runtime for the selection operator, because it requires the materialization of
the full feature vector. This materialization increases the runtime of selection by
up to 1.6$\times$ (Yelp, MLP small).

The best performance is achieved if the $\Delta$-representation and partial
evaluation of the classifier are used together. In this case, there is a
significant runtime speedup for both the mutation operator and selection
operators. Overall, this can lead to a performance improvement of
$5\times$ (Yelp, random forest).

\begin{figure}
  \centering

  \includegraphics[width=0.49\columnwidth]{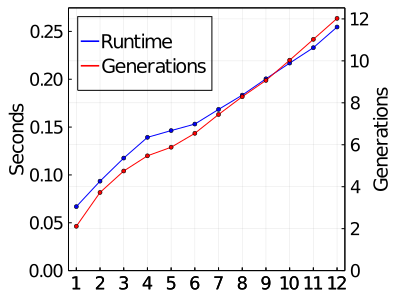}
  \includegraphics[width=0.49\columnwidth]{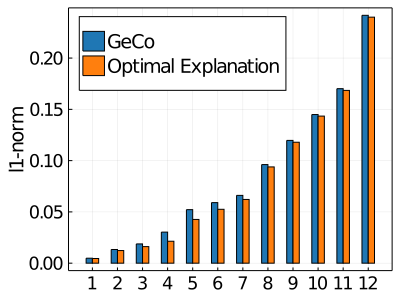}
  \vspace*{-1em}

  \caption{Evaluation of \geco over 100 Credit instances with synthetic
    classifiers that require 1-12 feature changes. (left) Average runtime and
    number of generations; (right) Average distance of \geco compared to the
    optimal explanation.}
  \label{fig:synclf}
\end{figure}

\mysubsubsec{Validating Explanation Quality} To evaluate whether \geco is
able to find good explanations, we design synthetic classifiers for which the
optimal explanation is known. Each classifier is a conjunction of unary
threshold conditions, and the outcome is positive iff all conditions are
satisfied. Given an instance that fails all conditions, the number of conditions
is equal to the number of features that the counterfactual needs to change. We
present further details on the synthetic classifiers in
Appendix~\ref{appendix:ground:truth}.

Figure~\ref{fig:synclf} presents the results of our evaluation on 100 Credit
instances which fail all conditions for all classifiers. We consider classifiers
with up to 12 conditions, which is the maximum number of features that can be
changed. We add features in the decreasing order of their domain sizes, which is
the most challenging order for \geco; we consider a different order in
Appendix~\ref{appendix:ground:truth}.

\geco always finds a valid counterfactual explanation, even if we require
changing all 12 features. The runtime is linear with respect to the number of
features changed, and proportional to number of generations of the genetic
algorithm. The distance of the explanations is always close to the distance of
the optimal explanation. In Appendix~\ref{appendix:ground:truth}, we show that,
by sampling more values during mutation, we can further decrease the distance
gap to the optimal explanation with minor performance degradation.

\mysubsubsec{Number of Generations and Explored Candidates}
Table~\ref{tbl:microbench} shows for each dataset how many generations \geco
needed on average to converge, and how many candidates it explored. The majority
(up to 97\%) of the candidates were generated by mutate.

\mysubsubsec{Compression by $\Delta$-representation} Table~\ref{tbl:microbench}
compares the sizes for the naive listing representation and the
$\Delta$-representation. We measure size in terms of the number of represented
values, and take the average over all generations for the  size of the candidate
population after the mutate and crossover operations. Overall, the
$\Delta$-representation can represent the candidate population up to $60\times$
more compactly than the naive listing representation.

\mysubsubsec{Effect of Constraints} We evaluate the impact of the
PLAF constraints on GeCo's runtime. The constraints without implications
significantly restrict the search space of feasible counterfactuals. Thus,
including these constraints improves the performance by 19.3\% for Adult and
32.2\% for Credit. The constraints with implications, however, may introduce
some overhead since they need to be checked dynamically using action cascading.
For example, the overhead is 19.2\% for Adult and 32.3\% for Credit.
Yet, the version with all constraints is still faster by 9.8\% for Credit
and 18.7\% for Adult over the version without any constraints. Finally, the
grouping of features also restricts the search space. For Adult, GeCo is
1.4$\times$ faster using the feature groups than without.

\ignore{
\subsection{Summary}

The  experiments show that \geco is the only system that can provide
high-quality explanations in real time. On the Credit and Adult datasets,  MACE
is the best competitor in terms of quality, but takes orders of magnitude longer
to run. The explanations for the fastest competitors (WIT and DiCE) have poor
quality.

The $\Delta$-representation and partial evaluation of the classifier can
significantly reduce \geco's end-to-end runtime. For our evaluation on the real
Allstate and Yelp datasets, the two optimizations improve the end-to-end runtime
by a factor of up to $5\times$.
}

%% file: include/sec7-conclusions.tex
\section{Conclusions}
\label{sec:conclusions}

We described \geco, the first interactive system for
counterfactual explanations that supports a complex, real-life
semantics of counterfactuals, yet provides answers in real time.
\geco\ defines a rich search space for counterfactuals, by considering
both a dataset of example instances, and a general-purpose constraint
language.  It uses a genetic algorithm to search for counterfactuals,
which is customized to favor counterfactuals that require the smallest
number of changes.  We described two powerful optimization techniques
that speed up the inner loop of the genetic algorithm:
$\Delta$-representation and partial evaluation.  We demonstrated
that, among five other systems reported in the literature, \geco\ is
the only one that can both compute quality explanations and find them
in real time.

\ignore{
}

This work opens up several directions for future work. First, counterfactual
explanations are subject to updates to the underlying data and classifier. We
plan to explore how we can generate explanations that are robust to small
changes in the data distribution or classifier. This is related to the more
general problem of robust machine learning. Second, \geco requires that the PLAF
constraints are provided by a domain expert. We plan to explore how we can
leverage constraints and dependencies in databases to generate these constraints
automatically. Third, \geco currently assumes that the input to the model is the
raw data. In practice, however, the model input is typically the output of
extensive feature engineering. We plan to explore how \geco can generate
explanations for the feature engineered data, but then return the corresponding
raw data values to the user. For structured relational data, the feature
engineering involves aggregating the raw data, in which case we would have to
connect \geco with techniques on explaining aggregate queries that have been
developed in the database community. Fourth, counterfactual explanations expose
values from the database, which may lead to privacy issues. We plan to explore
how to return explanation that satisfy both privacy and legislative
requirements. Finally, we have implemented partial evaluation manually, and it
is only supported for random forests and simple neural network classifiers.  We
plan to extend this optimization to other models by leveraging work from the
compilers community.

%% file: include/appendix.tex
\section{Additional Experiments}

In this section, we present further details and extensions for the experiments
presented in Sec.~\ref{sec:exp}.

\subsection{Constraints}
\label{appendix:constraints}

We present the PLAF constraints used in the experiments in Sec.~\ref{sec:exp}.
Figure~\ref{fig:constrains} presents the constraints we considered for the
Credit and Adult datasets.

For Adult, we enforce that \texttt{Age} and \texttt{Education} can only
increase, and \texttt{MaritalStatus}, \texttt{Relationship}, \texttt{Gender},
and \texttt{NativeCountry} cannot change. The implication specifies that if
\texttt{Education} increases then \texttt{Age} needs to increase as well. The
Adult dataset has two features that describe the education level,
\texttt{EducationNumber} and \texttt{EducationLevel}. Since these two features
are clearly correlated, we group them into a single feature group. For the
experiments in Sec.~\ref{sec:exp:end-to-end} we restrict the number of hours
worked per week to at most 60, since it is unrealistic to assume that the
considered instance will exceed this limit.

For Credit, we enforce that \texttt{AgeGroup}, \texttt{EducationLevel}, and
\texttt{HasHistoryOfOverduePayments} can only increase, while \texttt{isMale}
and \texttt{isMarried} cannot change. The first implication specifies that an
increase in education level for a teenaged customer results in an increase in
the AgeGroup from teenager to adult. The second implication specifies that if
the number of months with low spending in the last 6 months increases then the
number of months with high spending in the last 6 months should decrease.

For Yelp, we define a total of 18 constraints which specify that (1) the
location of a business cannot change (i.e., \texttt{city}, \texttt{state},
\texttt{longitude}, and \texttt{latitude} are immutable) and (2) past reviews
cannot be deleted (i.e., the review count or the number of compliments for
businesses cannot decrease). In addition, we synthesis two constraints with
implications. The first implication specifies that if the review count for a
business increases, then the business must be open. The second implication
states that if a review receives more ``cool'' tags, then we expect the business
to also receive more ``cool'' compliments.

\begin{figure}[t]
  \Description[]{}
  \setlength\tabcolsep{.5pt}
  \begin{tabular}{l}
    \multicolumn{1}{c}{{\bf Adult Dataset}}                               \\
    \texttt{PLAF x\_cf.Age >= x.Age}                                      \\
    \texttt{PLAF x\_cf.Education >= x.Education}                          \\
    \texttt{PLAF x\_cf.MaritalStatus = x.MaritalStatus}                   \\
    \texttt{PLAF x\_cf.Relationship = x.Relationship}                     \\
    \texttt{PLAF x\_cf.Gender = x.Gender}                                 \\
    \texttt{PLAF x\_cf.NativeCountry = x.NativeCountry}                   \\
    \texttt{PLAF IF x\_cf.Education > x.Education}                        \\
    \hspace{2.5em}\texttt{THEN x\_cf.Age >= x.Age+4}                      \\[1em]
    \multicolumn{1}{c}{{\bf Credit Dataset}}                              \\
    \texttt{PLAF x\_cf.isMale = x.isMale}                                 \\
    \texttt{PLAF x\_cf.isMarried = x.isMarried}                           \\
    \texttt{PLAF x\_cf.AgeGroup >= x.AgeGroup}                            \\
    \texttt{PLAF x\_cf.EducationLevel >= x.EducationLevel}                \\
    \texttt{PLAF x\_cf.HasHistoryOfOverduePayments >= }                   \\
    \hspace{2.5em}\texttt{x.HasHistoryOfOverduePayments}                  \\
    \texttt{PLAF x\_cf.TotalOverdueCounts >= x.TotalOverdueCounts}        \\
    \texttt{PLAF x\_cf.TotalMonthsOverdue >= x.TotalMonthsOverdue}        \\
    \texttt{PLAF IF x\_cf.EducationLevel > x.EducationLevel+1}            \\
    \hspace{2.5em}\texttt{\&\& x.AgeGroup < 2 THEN x\_cf.AgeGroup == 2}   \\
    \texttt{PLAF IF x\_cf.MonthsWithLowSpendingLast6Months >}             \\
    \hspace{2.5em}\texttt{x.MonthsWithLowSpendingLast6Months}             \\
    \hspace{2.5em}\texttt{THEN x\_cf.MonthsWithHighSpendingLast6Months <} \\
    \hspace{2.5em}\texttt{x.MonthsWithHighSpendingLast6Months}
  \end{tabular}
  \caption{PLAF constraints used for Adult and Credit.}
  \label{fig:constrains}
\end{figure}

\subsection{Number of Features Changed}
\label{appendix:numfeats}

\begin{figure}
  \centering
  \includegraphics[width=.8\columnwidth]{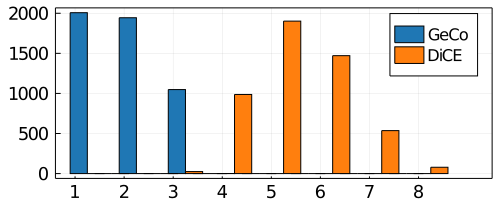}

  \caption{Distribution of number of features changed by \geco and DiCE on 5000 Adult instances.}
  \label{fig:exp1-dice2}
  \Description[]{}
\end{figure}

\begin{figure*}[tb]
  \centering

  \begin{tabular}{cc}
    {\bf Credit Dataset}                                                                         & {\bf Adult Dataset} \\
    \includegraphics[width=0.4\linewidth]{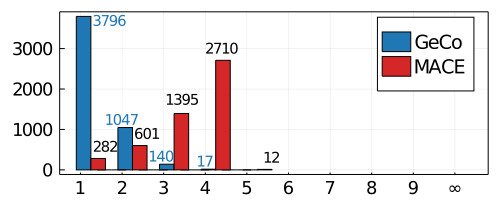}  &
    \includegraphics[width=0.4\linewidth]{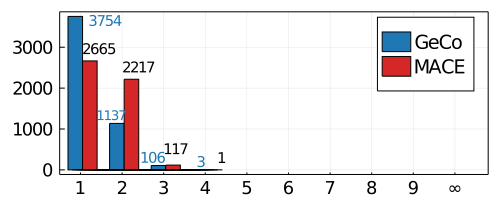}                         \\
    \includegraphics[width=0.4\linewidth]{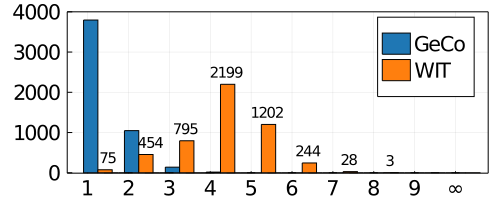}    &
    \includegraphics[width=0.4\linewidth]{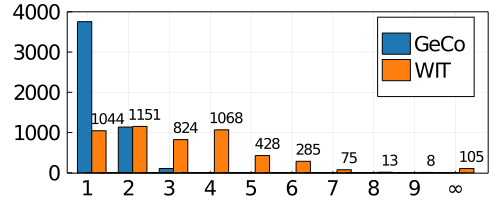}                           \\
    \includegraphics[width=0.4\linewidth]{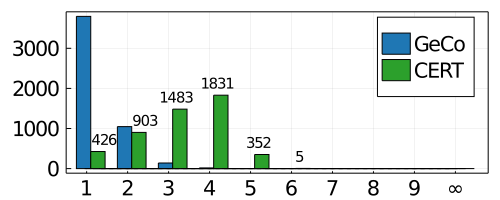}    &
    \includegraphics[width=0.4\linewidth]{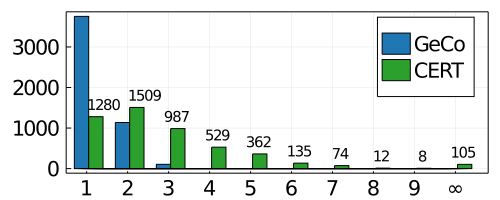}                           \\
    \includegraphics[width=0.4\linewidth]{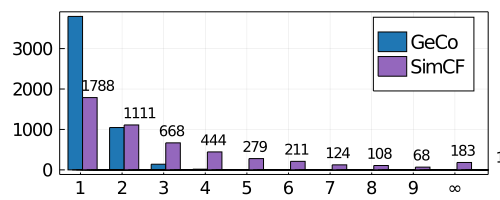} &
    \includegraphics[width=0.4\linewidth]{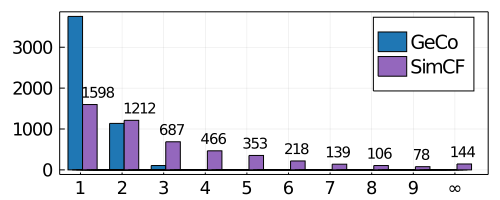}                        \\
  \end{tabular}






  \caption{Distribution of the number of features changed by \geco, MACE, WIT, CERT, and SimCF on Credit and Adult.}
  \label{fig:exp1-feat}
  \Description[]{}
\end{figure*}

The experiments in Sec.~\ref{sec:exp:tradeoff} showed that GeCo is able to find
explanations that are closer to the original instance than the five competing
systems. This can be attributed to the fact that GeCo is able to find
counterfactual explanations that require changing fewer features.
Figure~\ref{fig:exp1-feat} presents the distribution of the number of features
changed by MACE, WIT, CERT, and SimCF in comparison with the number of features
changed by \geco. For the Credit dataset, for instance, \geco can find
explanations with 1.27 changes on average, while MACE (the best competitor)
changes on average 3.39 features. WIT, CERT, and SimCF change on average 3.97,
3.15, and respectively 2.98 features.

Figure~\ref{fig:exp1-dice2} presents the corresponding distribution from the
comparison with DiCE. The plot shows again that \geco is able to find
explanations with few changes (1.8 changes on average), whereas DiCE changes on
average 5.4 features which is more than half of the total number of features
that can be changed.

\subsection{Synthetic Classifiers}
\label{appendix:ground:truth}

In this section, we present further details on the experiments with synthetic
classifiers for which we know the optimal explanation.

\mysubsubsec{Details on Classifiers}
Each classifier is a conjunction of unary threshold conditions (e.g.
\texttt{MostRecentBillAmount >= 4020.0}). This allows us to reason about the
optimal explanations for any instance that fails the threshold conditions. For
example, if the condition is \texttt{MostRecentBillAmount <= 4020.0}, then the
minimum change for  any instance that fails this condition would be to set
\texttt{MostRecentBillAmount = 4020.0}. The classifier returns 1 (the positive
outcome) if and only if all conditions are satisfied; otherwise we return
$0.5-d$, where $d$ denotes the normalized distance between the candidate and the
optimal explanation.

In Sec.\ref{sec:microbenchmarks}, we consider the following threshold conditions
for the mutable features of the Credit dataset:
\begin{enumerate}
  \item \texttt{MaxBillAmountOverLast6Months >= 4320.0}
  \item \texttt{MostRecentBillAmount >= 4020.0}
  \item \texttt{MaxPaymentAmountOverLast6Months >= 3050.0}
  \item \texttt{MostRecentPaymentAmount >= 1220.0}
  \item \texttt{TotalMonthsOverdue >= 12.0}
  \item \texttt{MonthsWithZeroBalanceOverLast6Months >= 1.0}
  \item \texttt{MonthsWithLowSpendingOverLast6Months >= 1.0}
  \item \texttt{MonthsWithHighSpendingOverLast6Months >= 3.0}
  \item \texttt{AgeGroup >= 2.0}
  \item \texttt{EducationLevel >= 3.0}
  \item \texttt{TotalOverdueCounts >= 1.0}
  \item \texttt{HasHistoryOfOverduePayments >= 1.0}
\end{enumerate}
We design 12 synthetic classifiers, where each classifier adds an additional
condition in the order shown above. For instance, the first classifier only
checks if \texttt{MaxBillAmountOverLast6Months >= 4320.0}, the second checks if
\texttt{MaxBillAmountOverLast6Months >= 4320.0 \&\& MostRecentBillAmount >= 4020.0},
and so on. The order in which the conditions are presented to the classifier is
the reverse order with respect to the size of their feature domains. This is the
most challenging order for \geco because it is more difficult to find the best
counterfactual when searching large domain spaces.

During the evaluation we consider only instances that fail all threshold
conditions in the classifier, which means that the counterfactual needs to
change at least as many features as there are conditions in the classifier.

\begin{figure*}[tb]
  \centering
  \includegraphics[width=.8\columnwidth]{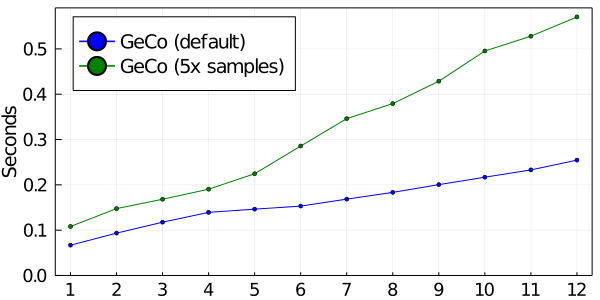}
  \hspace*{0.5em}
  \includegraphics[width=.8\columnwidth]{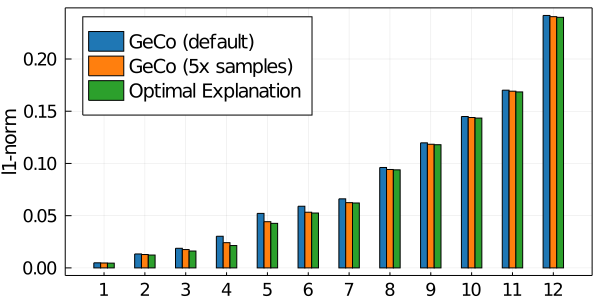}

  \caption{Comparison of two variants of GeCo using the synthetic classifiers over Credit. The first variant uses the default number of samples during mutation, and the second variant samples $5\times$ more values.}
  \label{fig:synclf-more-samples}
  \Description[]{}
\end{figure*}

\mysubsubsec{Evaluation with more samples} In Sec.\ref{sec:microbenchmarks}, we
show that, for the evaluation with synthetic classifiers, \geco is always able
to find a valid counterfactual explanation. However, as shown in
Fig.\ref{fig:synclf}, \geco may not always find the optimal explanation. This is
due to the fact that we are exploring large domain sizes, and the sampling of
action may not sample the optimal action for each feature. To mitigate the gap
in the distance between \geco's explanation and the optimal explanation we can
increase the number of samples during mutation.

The experiments in Fig.\ref{fig:synclf} assumed the default settings for the
number of samples during mutation $(\texttt{m}_{\text{init}}=20,
\texttt{m}_{\text{mut}}=5)$. In Fig.\ref{fig:synclf-more-samples} we compare two
variants of \geco. The first variant assumes the default settings, and the
second samples five times more values during mutation
$(\texttt{m}_{\text{init}}=100, \texttt{m}_{\text{mut}}=25)$. The results show
that the increase in the number of samples lowers the gap in distance between
\geco's explanation and the optimal explanation. This, however, comes with a
performance degradation since the increase in the number of samples requires
\geco to evaluate the classifier on more candidates. Even though we are sampling
$5\times$ more values, the performance is only $2\times$ slower for the
classifier that requires changing all twelve features.

\begin{figure}
  \centering
  \includegraphics[width=.49\columnwidth]{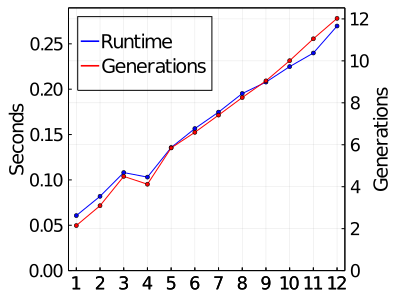}
  \includegraphics[width=.49\columnwidth]{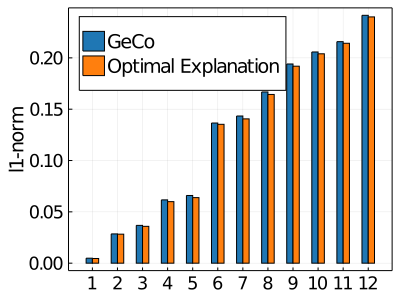}

  \caption{Evaluation of \geco over 100 Credit instances with the synthetic
    classifiers that require 1-12 feature changes. We add features one at a time
    using an order that interleaves features with large and small domain sizes.
    (left) Average runtime and number of generations; (right) Average distance of
    \geco compared to the optimal explanation.}
  \label{fig:synclf2}
  \Description[]{}
\end{figure}

\mysubsubsec{Evaluation with different feature order} In the previous sections,
we designed the synthetic classifiers following the decreasing order of the
feature domain sizes. Fig.\ref{fig:synclf2} presents the same evaluation using
the following order, where we interleave features with large and small domain
sizes (using the same threshold values):
\begin{enumerate}
  \item \texttt{MaxBillAmountOverLast6Months},
  \item \texttt{TotalOverdueCounts},
  \item \texttt{MostRecentBillAmount},
  \item \texttt{AgeGroup},
  \item \texttt{MaxPaymentAmountOverLast6Months},
  \item \texttt{HasHistoryOfOverduePayment},
  \item \texttt{MostRecentPaymentAmount},
  \item \texttt{TotalMonthsOverdue},
  \item \texttt{EducationLevel},
  \item \texttt{MonthsWithZeroBalanceOverLast6Months},
  \item \texttt{MonthsWithLowSpendingOverLast6Months},
  \item \texttt{MonthsWithHighSpendingOverLast6Months}.
\end{enumerate}

The results are comparable to Fig.\ref{fig:synclf}: the runtime depends to the
number of generations of the genetic algorithm, and the distance is consistently
close to one of the optimal explanation. The runtime plot, however, is not a
linear curve, because \geco finds explanations easier if we add a condition for
a feature with small domain size. For instance, adding \texttt{AgeGroup} in
experiment 4 has no significant effect on the runtime, because the dataset
contains only 5 age groups.

\subsection{Effect of Distance Function}
\label{appendix:distance}

Recall our distance function from Eq.~\eqref{eq:dist} and its parameters
$\alpha, \beta, \gamma$. In Sec.~\ref{sec:exp:tradeoff}, we show the explanation
quality for \geco on Credit and Adult using the parameters $(\alpha=0, \beta=1,
\gamma=0)$. When setting $(\alpha = 0.5, \beta = 0.5, \gamma = 0)$ and $(\alpha
= 0.33, \beta = 0.34, \gamma = 0.33)$, we observe that using the $\ell_0$-norm
($\alpha>0$) further reduces the number of features that \geco changes. In this
case, \geco always returns explanations that change only a single feature, for
both  Credit and Adult. This is because \geco can find explanations with a
single change, but the magnitude of this change may be larger than that of
changing multiple features. Thus, the user should only include the $\ell_0$-norm
if the goal is to limit the number of features changed. The $\ell_\infty$-norm
has no significant effect in our experiments, which may be because the
$\ell_1$-norm already restricts the maximum change.

We observed that including $\ell_0$-norm decreases the number of features
changed also in the experiments with synthetic classifiers. When we consider the
setting $(\alpha=0, \beta=1, \gamma=0)$, it is possible that \geco changes more
features than necessary. For instance, for experiments 1-4, \geco changes on
average 1.7, 2.9, 3.7, and respectively 4.4 features. When we use $(\alpha =
0.5, \beta = 0.5, \gamma = 0)$, however, \geco always changes only the number of
features that are required, because the $\ell_0$-norm enforces a penalty on
changing additional features. For the experiments that require five or more
feature changes, there is no difference in then number of features changed
between the two settings. In these experiments, \geco is required to run more
generations and thus automatically explores more combinations.